\documentclass[11pt]{article}

\pdfoutput=1

\usepackage[left=1.25in, top=1in, bottom=1in, right=1.25in]{geometry}

\usepackage{microtype}
\usepackage{graphicx}
\usepackage{subfigure}
\usepackage{booktabs} 

\usepackage[colorlinks=true,citecolor=blue]{hyperref}

\usepackage{booktabs} 
\usepackage{mathrsfs}
\usepackage{amssymb}
\usepackage{manfnt}  
\usepackage[framemethod=TikZ]{mdframed}  
\usepackage[inline]{enumitem}

\usepackage{natbib}
\usepackage{amsmath,amsthm}
\usepackage{stmaryrd}
\usepackage[mathscr]{eucal}
\usepackage{tcolorbox}

\newcommand{\argmin}{\operatornamewithlimits{argmin}}
\newcommand{\dd}{\mathrm{d}}
\newcommand{\indicator}[1]{\llbracket #1 \rrbracket}

\newcommand{\major}{{\,\succ_{{}_{\!\!\mathrm{M}}}\,}}
\newcommand{\unif}{{\mathrm{u}}}
\newcommand{\preclorenzeq}{{\,\preccurlyeq_{{}_{\!\!\!\!\!\!\!\mathrm{L}}}\,\,}}

\newcommand{\coloneqq}{:=}

\newcommand\independent{\protect\mathpalette{\protect\independenT}{\perp}}
\def\independenT#1#2{\mathrel{\rlap{$#1#2$}\mkern2mu{#1#2}}}

\newmdenv[backgroundcolor=red!12,
	linewidth=0.5pt,
	roundcorner=3pt,
	linecolor=red!20,
	skipabove=4pt,
	skipbelow=2pt,
        ]{summarybox}

\newtheorem{definition}{Definition}
\newtheorem{theorem}[definition]{Theorem}

\newtheorem{corollary}[definition]{Corollary}
\newtheorem{lemma}[definition]{Lemma}

\newtheoremstyle{theorem_noit}
  {\topsep}
  {\topsep}
  {}
  {0pt}
  {\bfseries}
  {.}
  { }
  {\thmname{#1}\thmnumber{ #2}\thmnote{ (#3)}}

\theoremstyle{theorem_noit}
\newtheorem{example}[definition]{Example}
\newtheorem{remark}[definition]{Remark}

\newcommand{\Bscr}{\mathscr{B}}

\newcommand{\Fscr}{\mathscr{F}}
\newcommand{\Lscr}{\mathscr{L}}
\newcommand{\Mscr}{\mathscr{M}}
\newcommand{\Pscr}{\mathscr{P}}

\newcommand{\Sscr}{\mathscr{S}}

\newcommand{\Ccal}{\mathcal{C}}
\newcommand{\Dcal}{\mathcal{D}}

\newcommand{\Fcal}{\mathcal{F}}
\newcommand{\Ical}{\mathcal{I}}
\newcommand{\Lcal}{\mathcal{L}}

\newcommand{\Rcal}{\mathcal{R}}

\newcommand{\Asf}{\mathsf{A}}
\newcommand{\Csf}{\mathsf{C}}
\newcommand{\Lsf}{\mathsf{L}}

\newcommand{\Ssf}{\mathsf{S}}
\newcommand{\Xsf}{\mathsf{X}}
\newcommand{\Ysf}{\mathsf{Y}}

\newcommand{\Zsf}{\mathsf{Z}}

\newcommand{\Ebb}{\mathbb{E}}
\newcommand{\Pbb}{\mathbb{P}}
\newcommand{\reals}{\mathbb{R}}
\newcommand{\naturals}{\mathbb{N}}

\newcommand{\CVaR}{\mathrm{CVaR}}

\newcommand{\todo}[1]{}

\newcommand{\DevM}{\Dcal}
\newcommand{\RiskM}{\Rcal}
\newcommand{\Risk}{L}
\newcommand{\SubRisk}[2]{\Risk_{#2}(#1)}
\newcommand{\SubRiskAbb}[1]{\Risk_{#1}}
\newcommand{\SubRiskRand}[1]{\Lsf(#1)}
\newcommand{\SubRiskRandAbb}{\Lsf}

\newcommand{\frm}{\mbox{\sc frm}}
\newcommand{\stim}{\mbox{\sc sim}}
\newcommand{\Rsim}{\Rcal_{\mbox{\scriptsize\sc sim}}}

\newcommand{\arxivVersion}[2]{#2}

\title{Fairness risk measures}
\author{
Robert C. Williamson \\
Australian National University \\
Canberra, Australia \\
\url{bob.williamson@anu.edu.au}
\and
Aditya Krishna Menon\thanks{Work done while at the Australian National University.} \\
Google \\
New York, USA \\
\url{adityakmenon@google.com}
}

\begin{document}
	\maketitle

\begin{abstract}


Ensuring that classifiers are non-discriminatory or \emph{fair} with respect to a sensitive feature (e.g., race or gender) is a topical problem.
Progress in this task requires fixing a definition of fairness,
and there have been several proposals in this regard over the past few years.
Several of these, however, assume either binary sensitive features (thus precluding categorical or real-valued sensitive groups), or result in non-convex objectives (thus adversely affecting the optimisation landscape).

In this paper, we propose a new definition of fairness that generalises some existing proposals, while allowing for generic sensitive features and resulting in a convex objective.
The key idea is to enforce that the expected losses (or \emph{risks}) across each subgroup induced by the sensitive feature are commensurate.
We show how this relates to the rich literature on \emph{risk measures} from {mathematical finance}.
As a special case, this leads to a new convex fairness-aware objective based on minimising the \emph{conditional value at risk} (CVaR).

\end{abstract}

\section{Introduction}


Ensuring that learned classifiers are non-discriminatory or \emph{fair} with respect to some sensitive feature (e.g., race or gender) is a topical problem~\citep{Pedreshi:2008,Zliobaite:2017,Chouldechova:2018}.
Progress on this problem requires that one agrees upon some pre-defined notion of fairness;
to this end, there have been several definitions of fairness at both the individual~\citep{Dwork:2012,Kusner:2017,Speicher:2018} and group level~\citep{Calders:2010,Feldman:2015,Hardt:2016,Zafar:2017,Heidari:2019}.

Recently, several works~\citep{Zafar:2017,Dwork:2018,Hashimoto:2018,Alabi:2018,Speicher:2018,Donini:2018,Heidari:2019} have abstracted earlier definitions of fairness by framing the problem in terms of \emph{subgroup losses}.
Intuitively, these works posit that a fair predictor incurs similar losses for each sensitive feature subgroup (e.g., men and women).
One encourages fairness by minimising specific notions of disparity of subgroup losses.
For specific choices of loss, this leads to a convex objective~\citep{Zafar:2017c,Donini:2018}.

In this paper, we propose a new definition of fairness that follows this theme,
but abstracts the notion of subgroup loss disparity.
Our resulting 
framework is applicable for
generic convex base losses (e.g., hinge),
complex sensitive features (e.g., multi-valued), 
and results in a convex objective.
In detail, our contributions are as follows:
\begin{itemize}[topsep=-2pt,itemsep=-4pt]
	\item[(\textbf{C1})] building on notions of fairness in terms of subgroup errors~\citep{Zafar:2017,Dwork:2018,Donini:2018},
	we provide a new definition of fairness
	(Definition~\ref{defn:d-fairness})
	requiring the average losses (or \emph{risks}) for each sensitive feature subgroup
	have low \emph{deviation}.
	
	\item[(\textbf{C2})] we draw a connection
	(Corollary~\ref{corr:frm-regular-coherent})
	between our proposed definition of fairness and 
	the rich literature on \emph{risk measures} from {mathematical finance}~\citep{Artzner:1999aa,Follmer:2011aa},
	thus allowing one to leverage tools and analyses from the latter.

	\item[(\textbf{C3})] we propose a new convex fairness-aware objective (Equation~\ref{eqn:cvar-fairness-variational})
	based on minimising the \emph{conditional value at risk} (CVaR)~\citep{Rockafellar:2000},
	and relate it to existing learning objectives.
\end{itemize}

In a nutshell, 
our proposal is to break up the standard risk into risks on each \emph{subgroup} defined by the sensitive feature.
We combine these
via an aggregator which
measures the mean \emph{and deviation} of the subgroup risks.
By defining some axioms an aggregator should satisfy, 
we obtain a connection to risk measures
from finance and operations research. 

We remark 
that much of 
the work in the paper
is in setting up the problem to easily exploit a wide body of existing results on risk measures; 
however, to our knowledge, the application of such tools to fairness is novel.
The end result is a simple, powerful framework to learn fair classifiers.

In the sequel, after reviewing existing work (\S\ref{subsec:formal-setup}),
we introduce our new definition of fairness (\S\ref{section:risk-theoretic}),
and relate it to financial risk measures 
(\S\ref{sec:axiomatic}).
We detail a special case employing the conditional value at risk (\S\ref{section:cvar}),
further develop
our approach
(\S\ref{section:implications}),
and confirm its empirical viability (\S\ref{section:experiments}).

\section{Background}
\label{subsec:formal-setup}


We briefly review the fairness-aware learning problem. 

\subsection{Standard and fairness-aware learning}

Given pairs of instances (e.g., job applicants) and target labels (e.g., likelihood of repaying a loan),
supervised learning concerns finding a predictor that best estimates the target label for new instances.
Formally, suppose there is a \emph{feature set} $X$, and \emph{label set} $Y$.
A \emph{predictor} is any $f \colon X \to A$ for some \emph{action set} $A$, where typically $A = Y$.
Suppose we are given a class of predictors $\Fscr \subseteq A^X$, and a \emph{loss function} $\ell \colon Y \times A \rightarrow \reals_{\ge 0}$ measuring the disagreement between a target label and its prediction. 
The
\emph{base goal} of learning is to find an $f^* \in \Fscr$ minimising the \emph{risk} or \emph{expected loss}:%
\footnote{We do not indicate the implicit dependence of $\Risk(f)$ on the underlying distribution or loss $\ell$ for brevity.}
\arxivVersion{
\begin{equation}
	\resizebox{0.9\linewidth}{!}{$\displaystyle
	f^* = \underset{f \in \Fscr}{\argmin} \, \Risk(f) \text{ where } \Risk(f) \coloneqq \underset{\Xsf,\Ysf}{\Ebb}\left[ \ell(\Ysf,f(\Xsf)) \right],%
	$}
	\label{eq:expected-risk-def}
\end{equation}
}
{
\begin{equation}
	f^* = \underset{f \in \Fscr}{\argmin} \, \Risk(f) \text{ where } \Risk(f) \coloneqq \underset{\Xsf,\Ysf}{\Ebb}\left[ \ell(\Ysf,f(\Xsf)) \right],%
	\label{eq:expected-risk-def}
\end{equation}
}
where $\Xsf,\Ysf$ are drawn from some distribution over $X \times Y$.

In fairness-aware learning, we augment the base goal by requiring our predictor does not discriminate with respect to some sensitive feature (e.g., race).
Formally, suppose there is a \emph{sensitive set} $S$
over which there is a random variable $\Ssf$,
and that the feature set $X$ contains $S$ as a subset.%
\footnote{Omitting $S$ from the feature set does not guarantee fairness, as it is typically correlated with other features~\citep{Pedreshi:2008}.}
A \emph{fairness measure} is some
$\Lambda( \cdot )$ for which $\Lambda( \Ysf, f( \Xsf ), \Ssf )$
evaluates the level of discrimination of $f$. 
The \emph{fairness goal} is to find an $f$ minimising the risk \emph{subject to} $\Lambda$ being small: for $\epsilon \geq 0$,
\begin{equation}
	f^* = \underset{f \in \Fscr}{\argmin} \, \Risk(f) \text{ such that } \Lambda( \Ysf, f( \Xsf ), \Ssf ) \leq \epsilon. %
	\label{eqn:fairness-aware-obj}
\end{equation}

\subsection{Measures of perfect fairness}
\label{sec:perfect-fairness}

To design a fairness measure $\Lambda$, it is useful to decide what it means for a predictor to be \emph{perfectly} fair.
Most formalisms of perfect (group) fairness are statements of statistical independence.
\emph{Demographic parity}~\citep{Dwork:2012} requires
\begin{equation}
	\label{eqn:dp}
	\Asf \independent \mathsf{S},
\end{equation}
so that knowledge of the predictions $\Asf \coloneqq f(\mathsf{X})$  provides no knowledge of
the sensitive feature $\mathsf{S}$.
For example, when $S = \{ {\tt male},
{\tt female} \}$, this would mean that the distribution of predictions are
identical for both men and women.
On the other hand, \emph{equalised odds}~\citep{Hardt:2016} requires
\begin{equation}
	\label{eqn:eo}
	\Asf \independent \mathsf{S} \mid \mathsf{Y},
\end{equation}
so that \emph{given} knowledge of the true label $\mathsf{Y}$, knowledge of
the predictions $\Asf$ provides no knowledge of the sensitive feature
$\mathsf{S}$.
Continuing the previous example, 
this requires that the
predictions do not discriminate between men and women \emph{beyond}
whatever power these have in predicting $\Ysf$.
Similarly, \emph{lack of disparate mistreatment}~\citep{Zafar:2017}
constrains the \emph{subgroup error rates} to be identical:
\arxivVersion{
\begin{equation}
	\label{eqn:ldm}
	\resizebox{0.9\linewidth}{!}{$\displaystyle
	( \forall s, s' \in S ) \, \mathbb{P}( \Ysf \neq \Asf \mid \Ssf = s ) = \mathbb{P}( \Ysf \neq \Asf \mid \Ssf = s' ).%
	$}
\end{equation}
}
{
\begin{equation}
	\label{eqn:ldm}
	( \forall s, s' \in S ) \, \mathbb{P}( \Ysf \neq \Asf \mid \Ssf = s ) = \mathbb{P}( \Ysf \neq \Asf \mid \Ssf = s' ).%
\end{equation}	
}
There are other extant notions of perfect fairness~\citep{Zafar:2017b,Ritov:2017,Heidari:2018,Zhang:2018},
including those for \emph{individual} rather than group fairness~\citep{Dwork:2012,Kusner:2017}.

\subsection{Measures of approximate fairness}
\label{sec:approx-fairness}

Notions of perfect fairness represent ideal statements about the world.
When learning a classifier from a finite training sample, it is infeasible to guarantee perfect fairness on a test sample~\citep{Agarwal:2018}.
In practice, one often instead works instead with measures of \emph{approximate fairness}.
The learner may then seek to achieve a \emph{tradeoff} between fairness and accuracy~\citep{Menon:2018}.

We highlight three popular measures of approximate fairness, using demographic parity (\ref{eqn:dp}) as the underlying perfect fairness notion for simplicity.
The first is to look at the maximal deviation between subgroup predictions~\citep{Calmon:2017}, \citep[Section 5.2.2]{Alabi:2018}:
\arxivVersion{
\begin{align*}
	\resizebox{0.99\linewidth}{!}{%
	$\displaystyle
	\Lambda_{\mathrm{dev}}( \Asf, \Ssf ) = \sup_{ a, s, s' } | \mathbb{P}( \Asf = a \mid \mathsf{S} = s ) - \mathbb{P}( \Asf = a \mid \mathsf{S} = s' ) |.
	$}%
\end{align*}
}
{
\begin{align*}
	\Lambda_{\mathrm{dev}}( \Asf, \Ssf ) = \sup_{ a, s, s' } | \mathbb{P}( \Asf = a \mid \mathsf{S} = s ) - \mathbb{P}( \Asf = a \mid \mathsf{S} = s' ) |.
\end{align*}
}
This measure is popular for binary $S$, where it is known as the \emph{mean difference score}~\citep{Calders:2010}.
However, it involves computing $|S|^2$ terms for categorical $S$, 
and is infeasible for real-valued $S$.
The former issue
can be addressed with a simple variant~\citep{Agarwal:2018}.

An elegant alternative is to recall that perfect fairness
measures assert that certain random variables are independent.
One may naturally measure approximate fairness by measuring their \emph{degree} of independence.
For example, one might quantify approximate demographic parity~(\ref{eqn:dp}) via
\begin{equation}
	\label{eqn:mi}
	\Lambda_{\mathrm{MI}}( \Asf, \Ssf ) = \mathrm{MI}( \Asf ; \mathsf{S} ) = \mathrm{KL}( \mathbb{P}( \Asf, \Ssf ) \, \| \, \mathbb{P}( \Asf ) \cdot \mathbb{P}( \Ssf ) ),
\end{equation}
where $\mathrm{MI}$ denotes the \emph{mutual information},
$\mathrm{KL}$ the Kullback-Leibler divergence,
$\mathbb{P}( \Asf, \Ssf )$ the joint distribution over predictions and sensitive features, and $\mathbb{P}( \Asf )$, $\mathbb{P}( \Ssf )$ the corresponding marginals.
Since 
the MI measures the degree of independence of two random variables,
$\Lambda_{\mathrm{MI}}$ is a natural measure of approximate demographic parity~\citep{Kamishima:2012,Fukuchi:2013,Calmon:2017,Ghassami:2018}.
One can replace the KL divergence in~(\ref{eqn:mi}) with
other measures of dissimilarity between distributions,
e.g., an $f$-divergence~\citep{Komiyama:2017} or
Hilbert-Schmidt criterion~\citep{Perez-Suay:2017}.

Conceptually, measures based on~(\ref{eqn:mi}) have appealing generality:
in particular, they can seamlessly handle multi-class, multi-label and continuous $S$.
However, they typically result in a non-convex objective~\citep{Kamishima:2012}.
An alternate measure
that is similarly general, 
but convex,
is the covariance between the target and sensitive features~\citep{Zafar:2017c,Olfat:2018,Donini:2018}:
\begin{equation}
	\label{eqn:cov}
	\Lambda_{\mathrm{cov}}( \Asf, \Ssf ) = \mathrm{Cov}( \Asf, \Ssf ) = \mathbb{E}[ \Asf \cdot \Ssf ] - \mathbb{E}[ \Asf ] \cdot \mathbb{E}[ \Ssf ].
\end{equation}

\subsection{Fairness-aware algorithms}
\label{sec:fair-algorithms}

Having fixed a notion of perfect or approximate fairness, one may then go about designing a fairness-aware learning algorithm.
Broadly, these follow one of three approaches:
\arxivVersion{
\begin{enumerate}[label=(\emph{\alph*}),itemsep=-2pt,topsep=-2pt]
	\item pre-process the training set to ensure fairness of \emph{any} learned model~\citep{Zemel:2013,Johndrow:2017,Calmon:2017,Adler:2018,delBarrio:2018,McNamara:2019};
	\item post-process model predictions to ensure their fairness~\citep{Feldman:2015,Hardt:2016};
	\item 
	directly ensure fairness by optimising (\ref{eqn:fairness-aware-obj})~\citep{Zafar:2016,Zafar:2017,Agarwal:2018,Donini:2018}.
\end{enumerate}
}
{
\begin{enumerate}[label=(\emph{\alph*}),itemsep=0pt,topsep=0pt]
	\item pre-process the training set to ensure fairness of \emph{any} learned model~\citep{Zemel:2013,Johndrow:2017,Calmon:2017,Adler:2018,delBarrio:2018,McNamara:2019};
	\item post-process model predictions to ensure their fairness~\citep{Feldman:2015,Hardt:2016};
	\item 
	directly ensure fairness by optimising (\ref{eqn:fairness-aware-obj})~\citep{Zafar:2016,Zafar:2017,Agarwal:2018,Donini:2018}.
\end{enumerate}
}
This paper focusses on methods of type (c);
we defer implications for methods of types (a) and (b) to future work.

\subsection{Scope of this paper}

In relation to the above (necessarily incomplete) survey, the scope of the present work is in 
providing:
\arxivVersion{
\begin{enumerate}[itemsep=-4pt,topsep=-4pt]
	\item[---] a new notion of approximate fairness (Definition~\ref{defn:d-fairness}),	
	\item[---] a new method that optimises for this notion (\S\ref{section:cvar}), and
	\item[---] a new connection between fairness and concepts from mathematical finance (Corollary~\ref{corr:frm-regular-coherent}).
\end{enumerate}
}
{
\begin{enumerate}[itemsep=0pt,topsep=0pt]
	\item[---] a new notion of approximate fairness (Definition~\ref{defn:d-fairness}),	
	\item[---] a new method that optimises for this notion (\S\ref{section:cvar}), and
	\item[---] a new connection between fairness and concepts from mathematical finance (Corollary~\ref{corr:frm-regular-coherent}).
\end{enumerate}
}
In more detail, we consider fairness in terms of \emph{subgroup risk}, following~\citep{Zafar:2017,Donini:2018,Dwork:2018,Alabi:2018}.
Our new notion of approximate fairness is that these risks exhibit low \emph{deviation}.
By connecting this to \emph{risk measures} in mathematical finance,
we arrive at a \emph{convex} objective for fairness-aware learning, applicable for generic sensitive features $S$, and with interesting connections to some existing learning paradigms.



\section{Fairness as subgroup risk deviation}
\label{section:risk-theoretic}


We present our new measure of fairness 
by introducing the notion of \emph{subgroup risks},
and using it to define natural measures of perfect (\S\ref{sec:perfect-subgroup}) and approximate fairness (\S\ref{sec:approx-subgroup}).
We also define some recurring notation, summarised in Table~\ref{tbl:glossary}.
The core idea of our proposal is to aggregate the subgroup risks by measuring their mean behaviour \emph{and} deviance (Equations~\ref{eqn:erm-agg} and~\ref{eqn:agg-risk}).

\begin{table}[!ht]
    \renewcommand{\arraystretch}{1.1} 
    \centering

    \begin{tabular}{llll}
        \toprule
        \toprule
        \textbf{Symbol} & \textbf{Meaning} \\
        \midrule
        $\ell, f$                   & Base loss, predictor \\
        $\Risk(f)$                  & Risk of $f$ on entire population \\
        $\SubRisk{f}{s}$            & Risk of $f$ on subgroup with $S = s$ \\
        $\SubRiskRand{f}$           & Random variable of all subgroup risks \\
        $\DevM( \SubRiskRand{f} )$  & Deviation of subgroup risks \\
        $\RiskM( \SubRiskRand{f} )$ & Aggregation of subgroup risks \\
        \bottomrule
    \end{tabular}

    \caption{Glossary of important symbols.}
    \label{tbl:glossary}
    \vspace{-\baselineskip}
\end{table}

\subsection{Subgroup risks}
\label{sec:subgroup-risk}

Observe that the sensitive feature $S$ partitions the instance space $X$ into subgroups (e.g., men and women).
It will be useful to define two induced quantities.
The first is the \emph{subgroup risk} for a predictor $f$, which for any $s \in S$ is
\begin{equation}
    \label{eqn:subgroup-risk}
    \SubRisk{f}{s} \coloneqq \underset{\Xsf, \Ysf |\Ssf=s}{\Ebb} \ell(\Ysf,f(\Xsf)). 
\end{equation}
The second is the random variable
$\SubRiskRand{f} \coloneqq \SubRisk{f}{\Ssf}$
summarising all subgroup risks.
For $|S| < \infty$, 
this is simply a discrete random variable taking on $|S|$ possible values, 
i.e.,
$\{ \SubRisk{f}{s} \}_{s \in S}$,
with 
corresponding probabilities
$\mathbb{P}( \Ssf = s )$.

We can now rewrite the original risk $\Risk(f)$ from (\ref{eq:expected-risk-def}) as an average over these subgroup risks:
\begin{equation}
    \label{eq:Rsf}
    \Risk(f) = \underset{\Ssf}{\Ebb} \, \underset{\Xsf, \Ysf \mid \Ssf}{\Ebb}[ \ell( \Ysf, f( \Xsf ) ) ] = \Ebb[ \SubRiskRand{f} ].%
\end{equation}
The base goal of learning (\ref{eq:expected-risk-def}) is thus expressible as
\begin{equation}
 \label{eqn:erm}
 f^* = \underset{f \in \Fscr}{\operatorname{argmin}} \, {\Ebb}[ \SubRiskRand{f} ],
\end{equation}
so that one seeks good average subgroup risk.
Equally, we wish to select $f^* \in \Fscr$ based on the expectations of the family of random variables $\{ \SubRiskRand{f} \}_{f \in \Fscr}$.

We now introduce our new measure of fairness.
Following the discussion in~\S\ref{subsec:formal-setup},
we do so in two steps:
we start by settling on a notion of perfect fairness based on the subgroup risks,
and then present an approximate version of the same.

\subsection{Perfect fairness via subgroup risks}
\label{sec:perfect-subgroup}

Every measure of fairness in \S\ref{sec:perfect-fairness} specifies that our predictor $f$ behaves similarly across the sub-groups induced by $S$.
We employ a notion of perfect fairness that is faithful to this.

\begin{definition}
\label{defn:perfect-utility}
We say that a predictor $f \in \Fscr$ is \emph{perfectly fair}
with respect to $\ell$
if
\emph{all subgroups attain the same average loss};
i.e.,
$\SubRiskRand{f}$
is a constant random variable, so that
\arxivVersion{
\begin{equation}
 \label{eqn:perfect-risk}
 \resizebox{0.875\linewidth}{!}{$\displaystyle
 ( \forall s, s' \in S ) \, \underset{\Xsf, \Ysf|\Ssf=s}{\Ebb} \ell(\Ysf,f(\Xsf)) = \underset{{\Xsf, \Ysf|\Ssf=s'}}{\Ebb} \ell(\Ysf,f(\Xsf)).%
 $}
\end{equation}}
{
\begin{equation}
 \label{eqn:perfect-risk}
 ( \forall s, s' \in S ) \, \underset{\Xsf, \Ysf|\Ssf=s}{\Ebb} \ell(\Ysf,f(\Xsf)) = \underset{{\Xsf, \Ysf|\Ssf=s'}}{\Ebb} \ell(\Ysf,f(\Xsf)).%
\end{equation}
}
\end{definition}

Abstractly, the idea behind (\ref{eqn:perfect-risk}) is that the loss $\ell$ 
should ideally be
chosen to
capture all aspects of the problem ignoring fairness;
perfect fairness means that regardless of the value of
sensitive attribute, the performance does not vary.
For a specific choice of $\ell$, Definition~\ref{defn:perfect-utility} captures an existing notion of perfect fairness owing to~\citet{Zafar:2017}.


%
\begin{example}
\label{ex:zafar}
For the 
zero-one loss
$\ell_{01}( y, f ) = \indicator{ y \neq f }$, (\ref{eqn:perfect-risk}) 
reduces to the previously introduced (\ref{eqn:ldm}):
\arxivVersion{
$$ \resizebox{0.99\linewidth}{!}{$\displaystyle( \forall s, s' \in S ) \, \mathbb{P}( f( \Xsf ) \neq \Ysf \mid \Ssf = s ) = \mathbb{P}( f( \Xsf ) \neq \Ysf \mid \Ssf = s' ).%
$}$$
}
{
$$ ( \forall s, s' \in S ) \, \mathbb{P}( f( \Xsf ) \neq \Ysf \mid \Ssf = s ) = \mathbb{P}( f( \Xsf ) \neq \Ysf \mid \Ssf = s' ).%
$$
}
\end{example}

Definition~\ref{defn:perfect-utility} is not new as a measure of perfect fairness.
Indeed,~\citet[Appendix H]{Donini:2018} considered essentially the same notion, with additional conditioning on $\Ysf = 1$.
Several other recent works implicitly define perfect fairness in terms of subgroup risks~\citep{Dwork:2018,Hashimoto:2018,Alabi:2018}.
Further, recent welfare-based notions of fairness~\citep{Speicher:2018,Heidari:2019} also posit that fair classifiers have equally distributed \emph{benefit} 
\arxivVersion{(i.e., negative losses; see Remark~\ref{rem:connection-welfare}).}
{(i.e., negative losses).}

However, we build on Definition~\ref{defn:perfect-utility} to provide a novel notion of \emph{approximate} fairness,
one which has appealing properties and provides a bridge to the tools of financial risk measures.

\subsection{Approximate fairness via subgroup deviations}
\label{sec:approx-subgroup}

A natural way to design an approximate fairness measure based on (\ref{eqn:perfect-risk})
is to 
ensure that the subgroup risks
$\SubRiskRand{f}$
are roughly constant.
Formally,
for some \emph{deviation measure} $\DevM$ of the non-constancy of a random variable
(e.g., the standard deviation),
we will require that 
$\DevM( \SubRiskRand{f} )$
is small.
\begin{definition}
\label{defn:d-fairness}
Let $\DevM( \cdot )$ be a measure of deviation of a random variable.
For any $\epsilon > 0$, we say that 
$f \in \Fscr$ is
$\epsilon$\emph{-approximately fair with respect to $\DevM$ and $\ell$} if
\emph{the average subgroup losses have small deviation};
i.e.,
$\DevM( \SubRiskRand{f} ) < \epsilon$.
\end{definition}

Definition~\ref{defn:d-fairness} is applicable for generic $S$ (e.g., real-valued).
For the case of binary $S$, it is consistent with existing notions of approximate fairness,
as we now illustrate.

\begin{example}
\label{ex:sd}
Suppose $S = \{0,1\}$,
and that we use deviation measure $\DevM_{\mathrm{SD}}( \cdot ) = \sigma( \cdot )$, where $\sigma$ is the standard deviation of a random variable.
Fix $f \in \Fscr$, and
for brevity write 
the subgroup risks as 
$\SubRiskAbb{s} \coloneqq \SubRisk{f}{s}$ and
$\SubRiskRandAbb \coloneqq \SubRiskRand{f}$.
We have
\begin{align}
    \DevM_{\mathrm{SD}}(\SubRiskRandAbb) &= \sqrt{\Ebb(\SubRiskRandAbb^2)-\Ebb^2(\SubRiskRandAbb)}
    = \frac{1}{2} \cdot |\SubRiskAbb{0}-\SubRiskAbb{1}|.\label{eq:SD-R-diff}
\end{align}
Recall that the subgroup risks $\SubRiskAbb{s}$ depend on the underlying loss $\ell$.
Employing the zero-one loss $\ell_{01}$ in (\ref{eq:SD-R-diff})
yields
\arxivVersion{
\resizebox{\linewidth}{!}{$\displaystyle
\DevM_{\mathrm{SD}}(\SubRiskRandAbb) = \frac{1}{2} \cdot | \Pbb(f(\Xsf)\ne\Ysf \mid \Ssf=0) - \Pbb(f(\Xsf)\ne\Ysf \mid \Ssf=1) |,
$}
}
{
$$ \DevM_{\mathrm{SD}}(\SubRiskRandAbb) = \frac{1}{2} \cdot | \Pbb(f(\Xsf)\ne\Ysf \mid \Ssf=0) - \Pbb(f(\Xsf)\ne\Ysf \mid \Ssf=1) |, $$
}

i.e., the mean-difference score~\citep{Calders:2010} applied to the 
lack of disparate mistreatment (\ref{eqn:ldm}).
\end{example}

\subsection{Fairness-aware learning via subgroup aggregation}

To achieve approximate fairness according to Definition~\ref{defn:d-fairness},
we may augment the standard expected risk~(\ref{eqn:erm}) 
with a penalty term: for suitable 
$\lambda > 0$, 
we may find
\begin{equation}
 \label{eqn:fair-erm}
 f^* = \argmin_{f\in\Fscr} \, \Risk(f) + \lambda \cdot \DevM( \SubRiskRand{f} ),
\end{equation}
so that we find a predictor that predicts the target label,
\emph{but}
does so consistently across all subgroups.
Observe now that in light of (\ref{eq:Rsf}),
we can succinctly summarise (\ref{eqn:fair-erm}) as
\begin{tcolorbox}[boxsep=0pt,left=2pt,right=2pt,top=-8pt,bottom=4pt,colback=black!5]
    \begin{align}
     \label{eqn:erm-agg}
     f^* &= \argmin_{f\in\Fscr} \RiskM_{\lambda}( \SubRiskRand{f} ) \\ 
     \label{eqn:agg-risk}
     \RiskM_{\lambda}( \SubRiskRandAbb ) &\coloneqq \Ebb( \SubRiskRandAbb ) + \lambda \cdot \DevM( \SubRiskRandAbb ).
    \end{align}
\end{tcolorbox}

We make two observations.
First,
both~standard risk minimisation (\ref{eqn:erm})
and ~(\ref{eqn:erm-agg})
minimise a function of the subgroup risks $\SubRiskRand{f}$;
the only difference is 
the choice of
\emph{subgroup risk aggregator} $\RiskM_{\lambda}$.
In~(\ref{eqn:erm-agg}), we aim to ensure that the subgroup risks are small,
\emph{and} that they are roughly commensurate.
Intuitively, the latter ensures that we do not exhibit systematic bias in terms of mispredictions on one of the subgroups.

Second,
given a finite sample $\lbag ( x_j, y_j, s_j ) \rbag_{j = 1}^m$, one may solve the empirical analogue of (\ref{eqn:erm-agg}):
we minimise $\RiskM_{\lambda}( \hat{\SubRiskRandAbb}(f) )$, where $\hat{\SubRiskRandAbb}$ comprises empirical subgroup risks,
i.e., we employ empirical expectations in (\ref{eqn:subgroup-risk});
see, e.g., ~(\ref{eqn:cvar-fairness-empirical}).

We make (\ref{eqn:erm-agg}) concrete with an example.

\begin{example}
For the setting of Example~\ref{ex:sd}, for deviation measure $\DevM_{\mathrm{SD}}$
we have
the fairness-aware objective (\ref{eqn:erm-agg}) 
\arxivVersion{
\begin{equation}
 \label{eqn:agg-sd-binary}
 \resizebox{0.88\linewidth}{!}{%
 $\displaystyle
 \RiskM_{\mathrm{SD}, \lambda}( \SubRiskRandAbb ) = \Ebb( \SubRiskRandAbb ) + \lambda \cdot \DevM_{\mathrm{SD}}( \SubRiskRandAbb ) = \Ebb( \SubRiskRandAbb ) + \frac{\lambda}{2} \cdot |\SubRiskAbb{1}-\SubRiskAbb{2}|,
 $}%
\end{equation}
}
{
\begin{equation}
 \label{eqn:agg-sd-binary}
 \RiskM_{\mathrm{SD}, \lambda}( \SubRiskRandAbb ) = \Ebb( \SubRiskRandAbb ) + \lambda \cdot \DevM_{\mathrm{SD}}( \SubRiskRandAbb ) = \Ebb( \SubRiskRandAbb ) + \frac{\lambda}{2} \cdot |\SubRiskAbb{1}-\SubRiskAbb{2}|,
\end{equation}
}
so that we ensure that the average subgroup risk is small,
\emph{and} that the two subgroup risks are commensurate.
\end{example}

\begin{remark}
\label{rem:beyond-unhinged}
For binary $S$,
previous methods sharing our notion of perfect fairness (Definition~\ref{defn:perfect-utility}) have objectives similar to (\ref{eqn:agg-sd-binary}).
There is, however, a subtle difference:
in (\ref{eqn:erm-agg}), we use the \emph{same} loss $\ell$ to measure the standard risk, and its deviation across subgroups.
However,~\citet{Zafar:2017,Donini:2018} employ \emph{different} losses for these two terms.
Specifically, they employ a linear loss for the deviation, which corresponds to measuring the covariance between $\Asf$ and $\Ssf$ per (\ref{eqn:cov}).
This choice is crucial to
ensuring convexity of their objective;
we shall see that one can preserve convexity for other $\ell$ by instead modifying $\DevM$.
\end{remark}

\begin{remark}
The idea of moving beyond expectations to a general aggregation of the \emph{per-instance} losses has
precedent in learning theory~\citep{Chapelle:2000,Maurer:2009}
and robust optimisation~\citep{Duchi:2016,Gotoh:2018}.
These encourage the loss deviance across \emph{all} samples to be small,
i.e.,
effectively, they treat each instance as its own group.
Similar connections will also arise in \S\ref{sec:relations}.
\end{remark}

\label{subsec:frm}

%

A natural question at this stage is what constitutes a ``sensible'' choice of deviation measure $\DevM$.
One may of course proceed with intuitively reasonable choices,
such as the standard deviation (Example \ref{ex:sd}); 
however, we shall now axiomatise the properties we would like \emph{any} sensible deviation measure to satisfy.
This shall lead to an admissible family of \emph{fairness risk measures}.

\section{Fairness risk measures}
\label{sec:axiomatic}

The proposal of the previous section was boiled down to a simple recipe
in (\ref{eqn:erm-agg}):
rather than minimise the average of the subgroup risks,
we minimise a general functional $\RiskM$ of them,
which involves an expectation \emph{and} deviation $\DevM$.
We now
axiomatically
specify the class of admissible subgroup aggregators $\RiskM$,
which will in turn specify the class of admissible deviations $\DevM$ (Theorem~\ref{theorem:quadrangle-first-part}).

The technical aspects here are not new;
rather, we leverage results in the risk measures literature
(particularly~\citet{Rockafellar:2013aa})
for a novel application to fairness.

\subsection{Fairness risk measures: an axiomatic definition}
\label{sec:axioms}

At this stage, we employ a slight change of terminology:
rather than refer to $\RiskM$ as a risk \emph{aggregator},
we shall refer to it as a risk \emph{measure}.
The reasoning for this change will become evident in the next section.

With this, we define the 
class of 
\emph{fairness risk measures}
$\RiskM$
as those satisfying
seven simple mathematical axioms.
In what follows, let $\Lcal^2(S)$ comprise real-valued random variables over $S$ with finite second moment.


\begin{definition}
\label{defn:frm}
We say $\RiskM\colon\Lcal^2(S)\rightarrow\bar{\reals} \coloneqq \reals \cup \{+\infty\}$ is a
\emph{fairness risk measure} if,
for any $\Zsf, \Zsf' \in \Lcal^2(S)$ and $C \in \reals$,
it satisfies 
the following
axioms (F1)--(F7):
\begin{description}[itemsep=-2pt,topsep=-2pt]
	\item[F1  Convexity]  $\RiskM( (1-\lambda)\Zsf + \lambda \Zsf') \le
		(1-\lambda) \RiskM(\Zsf) + \lambda\RiskM(\Zsf')$,
		$\forall\lambda\in(0,1)$.
	\item[F2  Positive Homogeneity] $\RiskM(0)=0$, $\RiskM(\lambda\Zsf)=\lambda\RiskM(\Zsf)$ 
		$\forall\lambda>0$.
	\item[F3 Monotonicity] $\RiskM(\Zsf)\le \RiskM(\Zsf')$ if
		$\Zsf\le\Zsf'$ almost surely.
	\item[F4 Lower Semicontinuity] 
		$\{\Zsf \colon \RiskM(\Zsf)\le C\}$
		is closed.
	\item[F5 Translation Invariance] 
		$\RiskM(\Zsf+C)= \RiskM(\Zsf) +C$.

	\item[F6 Aversity] $\RiskM(\Zsf)>\Ebb(\Zsf)$ for any non-constant random variable $\Zsf$.
	\item[F7 Law Invariance] 
		$\RiskM(\Zsf)=\RiskM(\Zsf')$
		if 
		$\Pbb_\Zsf = \Pbb_{\Zsf'}$.
\end{description}

\end{definition}


In Appendix~\ref{app:axiomatic-justification}, we argue why each of these axioms is natural when $\RiskM$ is used per (\ref{eqn:erm-agg}) to ensure fairness across subgroups.
Here, we highlight the import of two axioms:

	\emph{Convexity} (F1) is desirable because without it, the risk could be
	decreased by more fine grained partitioning as we now show.
	F1 and F2 are equivalent to $\RiskM$ being sub-additive and positive homogeneous~\citep{Rockafellar:2013aa}.
	Suppose $S=\{0,1\}$ and the sensitive feature is determinate and thus induces a partition $(X_0,X_1)$ of $X$.
	Then
	$\SubRiskRandAbb = \SubRiskRandAbb^{0} + \SubRiskRandAbb^{1}$,
	where $\SubRiskRandAbb^i$ is the restriction of $\SubRiskRandAbb$ to $X_i$,
	so that e.g.
	$\SubRiskRandAbb^0_s = \indicator{s = 0} \cdot \mathbb{P}( \Ssf = 0 ) \cdot \SubRiskRandAbb_s$.
	Now if $\RiskM$ were not convex, it would not be subadditive, and so
	$\RiskM(\SubRiskRandAbb^{0} + \SubRiskRandAbb^{1})=\RiskM(\SubRiskRandAbb) > \RiskM(\SubRiskRandAbb^{0})+\RiskM(\SubRiskRandAbb^{1})$.
	That is, by splitting into subgroups we could automatically make our risk measure smaller, which is counter to what we wish to achieve.
	
	Convexity is also desirable because, combined with F3, 
	if $f\mapsto \Risk(f)$ is convex, then so is $f\mapsto \RiskM(\SubRiskRand{f})$.
	Thus, 
	for convex $\ell$ and $\Fscr$,
	encouraging fairness does not pose an optimisation burden,
	in contrast to some existing approaches ~\citep{Kamishima:2012,Zafar:2016}.

	\emph{Aversity} (F6) has a clear justification, as it penalises deviation from perfect fairness (by Definition~\ref{defn:perfect-utility}, this corresponds to constant $\Lsf$);
	this is essential for any fairness measure.

\begin{remark}
The subgroup risk aggregator $\RiskM_{\mathrm{SD}}$ corresponding to the standard deviation (\ref{eqn:agg-sd-binary}) does not satisfy F1,
and thus is not a fairness risk measure.
This does not necessarily preclude its use;
while Appendix~\ref{app:axiomatic-justification} makes a case that these measures are sensible to use,
we do \emph{not} claim that these are the \emph{only} legitimate measures.
Nonetheless, we now see that a wide class of measures satisfy F1--F7.
\end{remark}


\subsection{Relation to financial risk measures}

In mathematical finance, a \emph{risk measure}~\citep{Artzner:1999aa} is a quantification of the potential loss associated with a position,
i.e.,
a function $\rho \colon \Lcal^2(S) \to \reals$ whose input is a random variable, being the possible outcomes for a position.
We now show the intimate relationship between fairness risk measures
and two classes of risk measures 
widely studied in
finance and operations research~\citep{Artzner:1999aa,Pflug:2007aa,Krokhmal:2011aa,Follmer:2011aa,Rockafellar:2013aa}. %
The first class is readily defined in terms of our existing axioms.
\begin{definition}
\label{defn:coherent}
We say $\RiskM \colon \Lcal^2(S) \rightarrow \bar{\reals}$
is a
\emph{coherent
measure of risk}~\citep{Artzner:1999aa} if it satisfies
F1 --- F5.
\end{definition}

The second class requires two additional axioms:
\begin{description}[itemsep=0pt,topsep=0pt]
	\item[F8 Translation Equivariance]  
$\RiskM(\Zsf)=C$ for any constant random variable $\Zsf$ taking value $C \in \reals$.
	\item[F9 Positivity under non-constancy]
	$\RiskM(\Zsf)\geq0$, with equality if and only if $\Zsf$ is constant.
\end{description}

Equipped with this, we have the following definition.

\begin{definition}
\label{defn:regular}
We say $\RiskM \colon \Lcal^2(S) \rightarrow \bar{\reals}$
is a \emph{regular measure of risk}~\citep{Rockafellar:2013aa}
if it satisfies 
F1, F4, F6 and F8.
Similarly, 
$\DevM \colon \Lcal^2(S) \rightarrow \bar{\reals}$
is a \emph{regular measure of deviation} if satisfies F1, F4 and F9.
\end{definition}

By employing $\Zsf=0$, (F5 $\land$ F6) $\implies$ F8.
This gives a simple relation between fairness and financial risk measures.

%
\begin{corollary}
\label{corr:frm-regular-coherent}
Every fairness risk measure is a coherent and regular measure of risk satisfying law-invariance.%
\footnote{Law-invariance is fortunately satisfied by most widely-used measures~
\citep{Rockafellar:2006aa, Pflug:2007aa}.}
\end{corollary}


%
\arxivVersion{
\begin{remark}
\label{rem:connection-welfare}
Our chosen axioms were inspired by \emph{risk} measures.
Recently,~\citet{Speicher:2018,Heidari:2019} proposed axioms
inspired by \emph{inequality} measures.
The two notions can be related;
see Appendix \ref{app:inequality}.
\end{remark}
}
{}

\subsection{Practical implications}
\label{subsec:optimization}

Connecting fairness and financial risk measures is not merely of conceptual interest.
In particular,
Corollary~\ref{corr:frm-regular-coherent} 
lets us construct fairness risk measures $\RiskM$
given a regular measure of deviation $\DevM$
via
$\RiskM( \Zsf ) = \Ebb( \Zsf ) + \DevM( \Zsf )$.
This is a consequence of
the following \emph{quadrangle theorem}.
\begin{theorem}[\citet{Rockafellar:2013aa}]
	\label{theorem:quadrangle-first-part}
	The relations 
	\begin{equation}
	 \label{eqn:quadrangle-theorem}
	 \RiskM(\Zsf)=\Ebb(\Zsf)+ \DevM(\Zsf) \text{ and } \DevM(\Zsf)=\RiskM(\Zsf)-\Ebb(\Zsf)
	\end{equation}
	give a one-to-one correspondence between regular measures
	of risk $\RiskM$ and regular measures of deviation $\DevM$. 
	Further, 
	$\RiskM$ is positively homogeneous iff $\DevM$ is positively homogeneous;
	and monotonic 
	iff $\DevM(\Zsf)\le \sup \Zsf - \Ebb(\Zsf)$ for all $\Zsf\in\Lcal^2(S)$.
\end{theorem}

\begin{remark}
\label{rem:looking-for-lambda}
Using the construction in (\ref{eqn:quadrangle-theorem}), we arrive at risk aggregators $\RiskM$ that are an expectation plus a deviance $\DevM$.
By contrast, in (\ref{eqn:fair-erm}) we applied a scalar $\lambda$ to the deviance.
This is equivalent to using a new deviance $\DevM_{\lambda} \coloneqq \lambda \cdot \DevM$.
\end{remark}

Corollary~\ref{corr:frm-regular-coherent}
also allows us to import well-studied financial risk measures for use in a fairness context,
as we now study.


\section{The CVaR-fairness risk measure}
\label{section:cvar}

We now illustrate a special case of our framework, where we use 
conditional value of risk (CVaR)
to measure subgroup deviation.
This is shown to yield a simple objective (Equation~\ref{eqn:cvar-fairness-variational}),
and connect to existing learning paradigms. 

%
\subsection{CVaR as a fairness risk measure}

We first recall the definition of CVaR.
For $\alpha\in(0,1)$ and random variable $\Zsf$,
let $q_\alpha( \Zsf )$ be the quantile at level $\alpha$.
The \emph{conditional value at risk} 
is~\citep{Rockafellar:2000}%
\footnote{%
We gloss over the subtleties of defining quantiles when $\Zsf$ has 
atomic components;
see \citep{Rockafellar:2013aa}.
} 
\begin{equation}
	\label{eqn:cvar}
	\CVaR_\alpha(\Zsf) \coloneqq \Ebb(\Zsf \mid \Zsf > q_\alpha(\Zsf)),
\end{equation}
i.e., it measures the tail behaviour of $\Zsf$.
Now define
		\begin{align}
			\label{eqn:agg-cvar}		
			\RiskM_{\mathrm{CV}, \alpha}( \Zsf ) &:= \mathrm{CVaR}_\alpha( \Zsf ) \\
			\DevM_{\mathrm{CV}, \alpha}( \Zsf )  &:= \mathrm{CVaR}_\alpha( \Zsf - \Ebb(\Zsf) ).
		\end{align}
Intuitively, $\DevM_{\mathrm{CV}, \alpha}$ measures 
the tail behaviour of $\Zsf' = \Zsf - \Ebb( \Zsf )$,
i.e.,
how much $\Zsf$ deviates above its mean.

One has that 
$\RiskM_{\mathrm{CV}, \alpha}$ 
and
$\DevM_{\mathrm{CV}, \alpha}$
are regular, coherent measures of risk and deviation respectively~\citep{Rockafellar:2013aa}.
By Theorem~\ref{theorem:quadrangle-first-part}, one may equally write 
$\DevM_{\mathrm{CV}, \alpha}( \Zsf ) = \mathrm{CVaR}_\alpha( \Zsf ) - \Ebb(\Zsf)$.
Further, $\RiskM_{\mathrm{CV}, \alpha}$ is a fairness risk measure with fairness-aware objective (\ref{eqn:erm-agg})
\arxivVersion{
\begin{tcolorbox}[boxsep=0pt,left=2pt,right=2pt,top=-4pt,bottom=2pt,colback=black!5]
	\begin{equation}
	 \label{eqn:cvar-frm}
	 \resizebox{0.88\linewidth}{!}{%
	 $\displaystyle
	 \min_{f \in \Fscr} \mathrm{CVaR}_{\alpha}( \SubRiskRand{f} ) = \min_{f \in \Fscr} \Ebb( \SubRiskRand{f} ) + \DevM_{\mathrm{CV}, \alpha}( \SubRiskRand{f} ).
	 $%
	 }%
	\end{equation}
\end{tcolorbox}
}
{
\begin{tcolorbox}[boxsep=0pt,left=2pt,right=2pt,top=2pt,bottom=2pt,colback=black!5]
	\begin{equation}
	 \label{eqn:cvar-frm}
	 \min_{f \in \Fscr} \mathrm{CVaR}_{\alpha}( \SubRiskRand{f} ) = \min_{f \in \Fscr} \Ebb( \SubRiskRand{f} ) + \DevM_{\mathrm{CV}, \alpha}( \SubRiskRand{f} ).
	\end{equation}
\end{tcolorbox}
}
Here, $\alpha \in (0, 1)$ is a tuning parameter.
From~(\ref{eqn:cvar}), increasing $\alpha$ focusses attention to the most extreme values of $\SubRiskRand{f}$, i.e., the largest subgroup risks.
Interestingly, the limiting cases of $\alpha$ yield famous fairness principles.
Per~\citet[Equation 5.8]{Rockafellar:2007},
as $\alpha \to 1$, (\ref{eqn:cvar-frm}) becomes
\begin{equation}
	\label{eqn:min-max}
	\min_{f \in \Fscr} \max_{s \in S} \SubRisk{f}{s}, 
\end{equation}
i.e., we seek \emph{all} subgroup risks to be small,
per the maximin principle~\citep{Rawls:1971aa}.
As $\alpha \to 0$, (\ref{eqn:cvar-frm}) becomes
$$ \min_{f \in \Fscr} \Ebb_{\Ssf}( \SubRisk{f}{\Ssf} ), $$
i.e., we seek the \emph{average} subgroup risks to be small,
per the impartial observer principle~\citep{Harsanyi:1977ac}
for uniform $\Ssf$ (see~\S\ref{sec:weighting}).
%
To intuit the effect of generic $\alpha \in (0, 1)$,
suppose 
$n = |S| < \infty$,
and $\Ssf$ has uniform distribution.
Then, 
\arxivVersion{
\begin{equation}
 \label{eqn:cvar-quantile-general}
 \resizebox{0.88\linewidth}{!}{%
 $\displaystyle
 \mathrm{CVaR}_{\alpha}( \SubRiskRand{f} ) = \frac{\lambda_\alpha}{k_\alpha} \sum_{i = 1}^{k_\alpha} \Risk_{[i]}( f ) + (1 - \lambda_\alpha) \cdot \Risk_{[k_\alpha+1]}( f ),
 $
 }
\end{equation}
}
{
\begin{equation}
 \label{eqn:cvar-quantile-general}
 \mathrm{CVaR}_{\alpha}( \SubRiskRand{f} ) = \frac{\lambda_\alpha}{k_\alpha} \sum_{i = 1}^{k_\alpha} \Risk_{[i]}( f ) + (1 - \lambda_\alpha) \cdot \Risk_{[k_\alpha+1]}( f ),
\end{equation}
}
where $\Risk_{[i]}( f )$ denotes the $i$th largest subgroup risk,
$k_\alpha \coloneqq \lceil n \alpha \rceil$ and 
$\lambda_\alpha$ is a weighting parameter given by~\citet[Proposition 8]{Rockafellar:2002}.
When 
$k_\alpha$ is an integer, 
\begin{equation}
 \label{eqn:cvar-quantile}
 \mathrm{CVaR}_{\alpha}( \SubRiskRand{f} ) = \frac{1}{k_\alpha} \sum_{i = 1}^{k_\alpha} \Risk_{[i]}( f ), 
\end{equation}
Minimising (\ref{eqn:cvar-frm}) 
seeks that
the average of the \emph{largest} subgroup risks is small.
This tightens the range of subgroup risks,
thus ensuring they are commensurate.
%
Thus, using CVaR as an aggregator (or deviance measure) yields intuitive objectives.
We now show these are feasible to optimise.

\begin{remark}
The maximal subgroup risk (\ref{eqn:min-max}) was also considered in~\citet{Hashimoto:2018},
motivated by settings where group identity is \emph{unknown}.
Objectives that interpolate between maximum and average subgroup risk have been proposed, e.g.,~\citet[Section 6.1]{Alabi:2018}.
These are similar in spirit to (\ref{eqn:cvar-quantile-general}); 
note however that
(\ref{eqn:cvar-quantile-general}) allows one to choose any $\alpha \in (0, 1)$, and thus effectively account for a \emph{partial} version of the $(k_\alpha+1)$th largest subgroup risk.
\end{remark}

\begin{remark}
Increasing both
$\lambda \in (0,\infty)$ and $\alpha \in (0,1)$ in the fairness risk measures
$\Rcal_{\mathrm{SD},\lambda}$ (Equation \ref{eqn:agg-sd-binary}) and
$\Rcal_{\mathrm{CV},\alpha}$ (Equation \ref{eqn:agg-cvar})
penalise variability of subgroup risks $\SubRiskRand{f}$. 
But the effects are different in detail: large $\lambda$ means strong control on the variance of $\SubRiskRand{f}$, whereas large $\alpha$ means that attention is only paid to the most extreme values of $\SubRiskRand{f}$.
\end{remark}

\subsection{Optimising CVaR-fairness}

Optimisation of quantities based on the CVaR is aided by a \emph{variational representation}:
for any $\alpha \in (0,1)$ and random variable $\Zsf$,~\citep[Theorem 1]{Rockafellar:2000}
\begin{equation}
 \label{eqn:cvar-variational}
 \mathrm{CVaR}_\alpha( \Zsf ) = \min_{\rho \in \reals} \left\{ \rho + \frac{1}{1 - \alpha} \cdot \Ebb[ \Zsf - \rho ]_+ \right\}. 
\end{equation}
Consequently,
the CVaR-fairness objective (\ref{eqn:cvar-frm}) becomes 
\arxivVersion{
\begin{tcolorbox}[boxsep=0pt,left=2pt,right=2pt,top=6pt,bottom=6pt,colback=black!5]
	\begin{equation}
	 \label{eqn:cvar-fairness-variational}
	 \min_{f \in \Fscr, \rho \in \reals} \left\{ \rho + \frac{1}{1 - \alpha} \cdot \Ebb[ \SubRiskRand{f} - \rho ]_+ \right\}.
	\end{equation}
\end{tcolorbox}
}
{
\begin{tcolorbox}[boxsep=0pt,left=2pt,right=2pt,top=6pt,bottom=6pt,colback=black!5]
	\begin{equation}
	 \label{eqn:cvar-fairness-variational}
	 \min_{f \in \Fscr, \rho \in \reals} \left\{ \rho + \frac{1}{1 - \alpha} \cdot \Ebb[ \SubRiskRand{f} - \rho ]_+ \right\}.
	\end{equation}
\end{tcolorbox}
}
This is a convex objective
when $f \mapsto \SubRiskRand{f}$ is convex
(e.g., using a convex base $\ell$ and $\Fscr$).
Given a finite sample $\lbag ( x_j, y_j, s_j ) \rbag_{j = 1}^{m}$
with $n = |S| < +\infty$,
this becomes
\arxivVersion{
\begin{equation}
 \label{eqn:cvar-fairness-empirical}
 \resizebox{0.88\linewidth}{!}{$\displaystyle
 \min_{f \in \Fscr, \rho \in \reals} \left\{ \rho + \frac{1}{n \cdot (1 - \alpha)} \sum_{s \in S} \left[ \frac{1}{m_s} \sum_{j \colon s_j = s} \ell( y_j, f( x_j ) ) - \rho \right]_+ \right\},$%
 }
\end{equation}
}
{
\begin{equation}
 \label{eqn:cvar-fairness-empirical}
 \min_{f \in \Fscr, \rho \in \reals} \left\{ \rho + \frac{1}{n \cdot (1 - \alpha)} \sum_{s \in S} \left[ \frac{1}{m_s} \sum_{j \colon s_j = s} \ell( y_j, f( x_j ) ) - \rho \right]_+ \right\},%
\end{equation}
}
for $m_s$ the number of examples with sensitive feature $s$.
In words, for fixed $\rho$, we find a predictor $f \in \Fscr$ which minimises a variant of the standard expected risk,
wherein 
we discard all subgroup risks which are smaller than $\rho$;
i.e., 
we focus attention on the ``hard'' subgroups.

%
\subsection{Relation to existing paradigms}
\label{sec:relations}

\citet{Fan:2017} considered the problem of learning 
a
robust binary classifier
given a sample $\lbag ( x_j, y_j ) \rbag_{j = 1}^m$
and loss $\ell \colon Y \times A \to \reals$.
To achieve this, it was proposed to
minimise the average of the \emph{top-$k$ per-instance} losses for $k \ll m$:
\begin{equation}
	\label{eqn:top-k}
	\min_{f \in \Fscr} \frac{1}{k} \sum_{i = 1}^{k} \ell_{[i]}( f ), 
\end{equation}
where $\ell_{[i]}( f )$ is the $i$th largest element of the per-instance losses
$[ \, \ell( y_j, f( x_j ) ) \, ]_{j = 1}^m$.
Following (\ref{eqn:cvar-quantile}), this is equal to%
\footnote{The connection to CVaR was not explicitly noted in~\citet{Fan:2017}.
However, they employed the variational representation (\ref{eqn:cvar-variational})
as
derived in a different context by~\citet{Ogryczak:2003}.}
$$ \min_{f \in \Fscr} \mathrm{CVaR}_{\alpha_k}( \SubRiskRandAbb_{\mathrm{inst}}(f) ) $$
where 
${\alpha_k} \coloneqq {k}/{m}$,
and
$\SubRiskRandAbb_{\mathrm{inst}}(f)$ is the discrete random variable of \emph{per-instance losses}, with values $\{ \ell( y_j, f( x_j ) \}_{j = 1}^m$.
Consequently,
despite being developed with a wholly different goal in mind,
this objective is a special case of our framework where \emph{each instance belongs to a separate group}.

CVaR also arises in the $\nu$-SVM~\citep{Scholkopf:2000},
which alternately parametrises the SVM
with 
$\nu \in (0, 1)$, and
whose objective is expressible as~\citep{Gotoh:2005,Takeda:2008,Tsyurmasto:2014}
$$ \min_{f \in \mathscr{H}} \frac{1}{2} \| f \|_{\mathscr{H}}^2 + \nu \cdot \mathrm{CVaR}_{1 - \nu}( \mathsf{M}(f) ), $$
where $\mathsf{M}(f)$ is the random variable of \emph{per-instance margins},
taking values $[ \, -y_j \cdot f( x_j ) \, ]_{j = 1}^m$.
This is a special case of our framework where {each instance belongs to a separate group},
and one employs the ``linear'' loss $\ell( y, f ) = -y \cdot f$:
while the $\nu$-SVM ignores (or down-weights) any \emph{instance} with low \emph{margin error},
we ignore (or down-weight) any \emph{subgroup} with low \emph{average loss}.


\section{Extensions and discussion}
\label{section:implications}

We briefly observe some extensions of our formulation.

\subsection{Sensitive feature weighting}
\label{sec:weighting}

In forming our fairness-aware objective (\ref{eqn:erm-agg}),
we employed the standard risk $\Risk(f)$, which is a weighted sum of the subgroup risks (Equation~\ref{eq:Rsf}).
The default weighting is the underlying sensitive feature distribution.
However, one could easily apply different a weighting $\nu_S$ to privilege certain groups over others.
For $|S| < \infty$, we could define 
(c.f.~(\ref{eq:Rsf}))
\begin{equation}
 \label{eqn:weighted-risk}
 \Risk(f;\nu_S) \coloneqq \underset{\Ssf \sim \nu_S}{\Ebb}[ \SubRisk{f}{\Ssf} ] = \sum_{s \in S} \nu_S( s ) \cdot \SubRisk{f}{i}. 
\end{equation}
For example, when $S = \{0,1\}$, if one felt that individuals with $s=0$ were more important to treat well, one could simply put a large mass on $1$, e.g. $\nu_S( 0 )=0.9$ and $\nu_S( 1 ) = 0.1$.
The effects of imposing $\Ssf \sim \nu_S$ will similarly be reflected in one's deviation measure $\DevM( \SubRisk{f}{\Ssf} )$.

To treat both groups equally in terms of risk, one could alternately choose $\nu_S$ to be uniform.
This forms the basis for Harsanyi's principle of justice~\citep{Harsanyi:1977ac},
and would be analogous to the use of the balanced error in classification~\citep{Brodersen:2010,Menon:2013}.

\begin{figure*}[!t]
	\centering
	\includegraphics[scale=0.28]{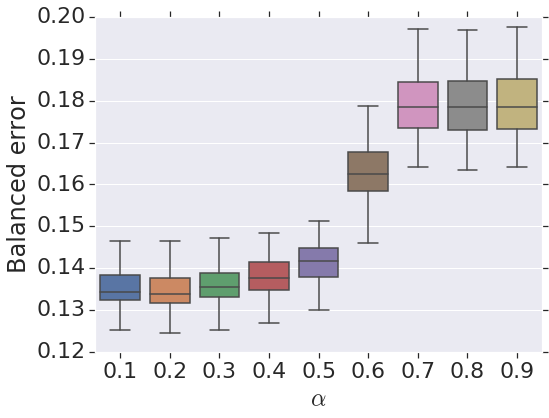}
	\quad
	\includegraphics[scale=0.28]{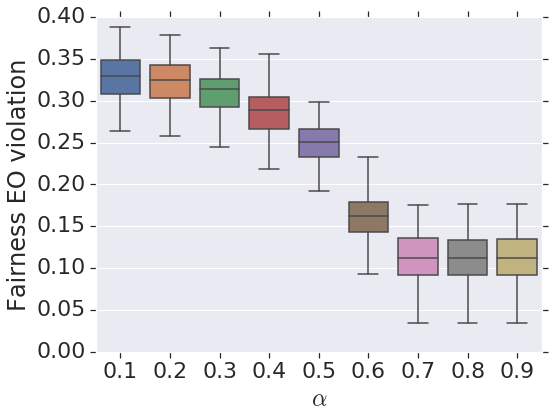}
	\quad
	\includegraphics[scale=0.28]{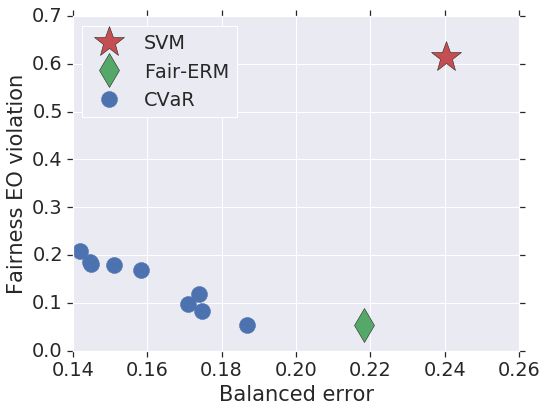}
	\quad
	\includegraphics[scale=0.27]{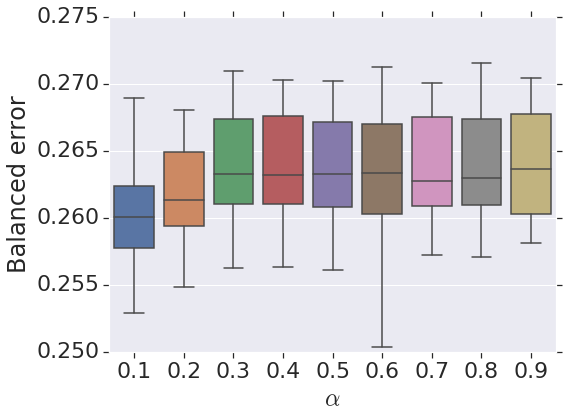}
	\quad
	\includegraphics[scale=0.27]{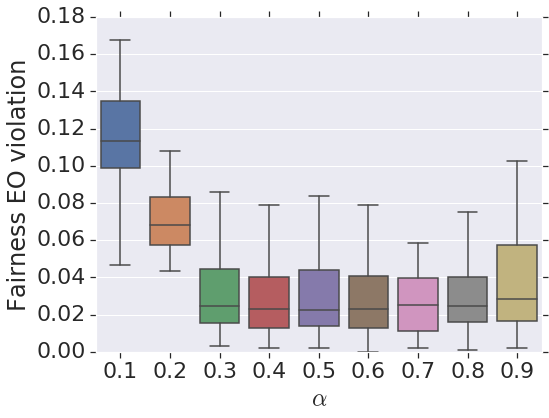}
	\quad
	\includegraphics[scale=0.27]{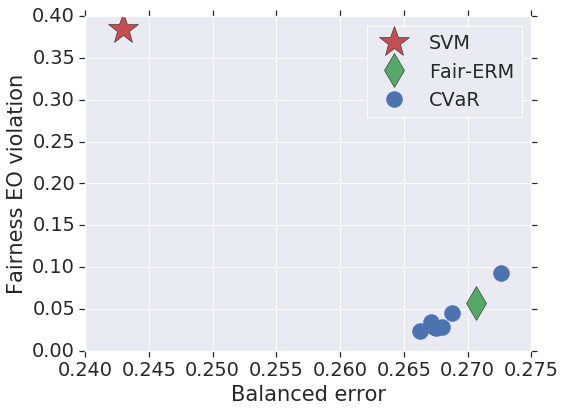}	
	\caption{Results on {\tt synth} (top) and {\tt adult} (bottom) datasets.
	The left and middle panel 
	show that as $\alpha$ is varied,
	CVaR-based optimisation results in a decrease in predictive accuracy and fairness violation.
	The right panel overlays the performance of CVaR-based optimisation for various $\alpha$ with that of two baselines.
	The CVaR method is shown to attain a reasonable fairness-accuracy tradeoff.
	}
	\label{fig:cvar-synth-alpha}
\end{figure*}

\subsection{Non-binary sensitive features}
\label{subsec:fine-graining}

Our examples thus far have focussed on binary $S$.
However, 
the risk measures underpinning our framework seamlessly handle generic $S$. 
We make this concrete with two examples.
The first is where $S=\reals_{\ge 0}$ (as is appropriate for a person's income, e.g.).
Then,
for $\alpha \in (0, 1)$ 
and
measure $\nu_S$ over $S$ per~(\ref{eqn:weighted-risk}),
the CVaR-fairness
objective (\ref{eqn:cvar-fairness-variational}) is: 
\begin{equation}
 \label{eqn:cvar-continuous}
 \min_{f \in \Fscr, \rho \in \reals} \left\{ \rho + \frac{1}{1 - \alpha} \cdot \int_S [ \SubRisk{f}{s} - \rho ]_+ \, \nu_S( \mathrm{d}s ) \right\}. 
\end{equation}
On a finite sample $\lbag ( x_j, y_j, s_j ) \rbag_{j = 1}^m$ with all $s_j$'s distinct for simplicity,
taking the empirical measure $\hat{\nu}_S$ gives 
\begin{equation}
	\label{eqn:cvar-continuous-empirical}
	\min_{f \in \Fscr, \rho \in \reals} \left\{ \rho + \frac{1}{1 - \alpha} \cdot \frac{1}{m} \sum_{j = 1}^m \left[ \ell( y_j, f( x_j ) ) - \rho \right]_+ \right\}, 
\end{equation}
so that each instance is considered as belonging to the same group.
Interestingly, this is equivalent to the top-$k$ objective (\ref{eqn:top-k}) for $k = m \alpha$.
However, one may consider other natural alternatives;
e.g., one may construct a non-parametric estimate of $\nu_S$ from the given sample, and use this in (\ref{eqn:cvar-continuous}).


The case of multiple sensitive features $\{ S_1, \ldots, S_l \}$ can be similarly handled:
all one needs to do is define a suitably structured $S$, and a valid measure $\nu_S$ over $S$.
As an example, 
one can set $S := S_1\times\cdots\times S_l$ and define 
$\nu_S$ as the product of measures $\nu_{S_i}$ on each individual sensitive feature.

\section{Empirical illustration}
\label{section:experiments}

We present experiments illustrating the practical viability of our framework.
In particular, we demonstrate that optimising the CVaR-fairness objective (\ref{eqn:cvar-fairness-empirical})
yields solutions with reasonable fairness-accuracy tradeoffs.

%


In detail,
we assess the performance of CVaR-based optimisation (\ref{eqn:cvar-fairness-empirical}) as $\alpha$ is tuned in $\{ 0.1, 0.2, \ldots, 0.9 \}$.
As baselines, we compare against a standard SVM,
and the fair-ERM approach of~\citet{Donini:2018}.
For all methods, we use square-hinge $\ell( y, f ) = [ 1 - y f ]_+^2$ as our base loss,
and use regularised linear scorers as our $\Fscr$. 
We use the validation procedure of~\citet{Donini:2018},
to tune the regularisation strength,
but with balanced in place of 0-1 error as the primary measure of predictive performance on $\Ysf$. 

We present results
on a
synthetic two-dimensional dataset ({\tt synth}) 
from~\citet{Donini:2018},
and the UCI {\tt adult} dataset
with gender as the binary $S$.
(In Appendix~\ref{app:experiments}, we present additional results, including on a real-valued $S$.)
For each method, 
over 100 random 80---20\% train-test splits
we measure the predictive performance on $\Ysf$ via the balanced error,
and the fairness on $\Ssf$ via the violation of equality of opportunity (EO)~\citep{Hardt:2016}, as measured by $| \mathbb{P}( \Asf = 1 \mid \Ssf = 1, \Ysf = 1 ) - \mathbb{P}( \Asf = 1 \mid \Ssf = 0, \Ysf = 1 ) |$.
As both datasets are slightly imbalanced, we weight positives and negatives equally for each method.

The left and middle panels of 
Figure~\ref{fig:cvar-synth-alpha} evince that as $\alpha$ is increased,
there is a decrease in predictive accuracy accompanied by an increase in fairness (i.e., decreased violation of the EO condition).
We remark here that the CVaR method only explicitly encourages the violation with respect to the square-hinge loss is minimised across the subgroups, 
which is indeed manifest (see Appendix~\ref{app:experiments}).

The right panel of 
Figure~\ref{fig:cvar-synth-alpha} summarises the fairness-accuracy tradeoff for all methods on one train-test split.
For the CVaR method, we present operating points for all values of $\alpha$.
Generally, CVaR's results are competitive with the fair-ERM approach of~\citet{Donini:2018}.
While more extensive experiments are apposite,
the above indicates the practical promise in further studying fairness risk measures.

\section{Conclusion and future work}
\label{section:conclusion}

We proposed a new definition of fairness that generalises some existing proposals, while allowing for generic sensitive features and resulting in a convex objective.
The key idea is to enforce that the expected losses (or \emph{risks}) across each subgroup induced by the sensitive feature are commensurate.
We showed how this relates to the rich literature on \emph{risk measures} from {mathematical finance}.
As a special case, this leads to a new convex fairness-aware objective based on minimising the \emph{conditional value at risk} (CVaR).

Our relating of fairness and risk measures motivates
study of
risk measures beyond CVaR,
e.g., 
spectral measures~\citep{Acerbi:2002aa},
optimised certainty equivalents~\citep{Ben-Tal:2007},
\&
entropic value at risk~\citep{Ahmadi-Javid:2012}.




\arxivVersion{
\bibliographystyle{icml2019}
}
{
\bibliographystyle{plainnat}
}
\bibliography{references}

\clearpage

\appendix
\onecolumn

\begin{center}
  {\LARGE\bf Supplementary material for ``Fairness risk measures''}
\end{center}



\section{Justification of fairness risk measure axioms}
\label{app:axiomatic-justification}

We now argue why each of these axioms is natural when $\RiskM$ is used per (\ref{eqn:erm-agg}) to ensure fairness across subgroups.
We note that apart from (F2),
none of these properties can be relaxed without causing problems.

\emph{Convexity} (F1) is desirable because without it, the risk could be
decreased by more fine grained partitioning as we now show.
F1 and F2 are equivalent to $\RiskM$ being sub-additive and positive homogeneous~\citep{Rockafellar:2013aa}.
Suppose $S=\{0,1\}$ and the sensitive feature is determinate and thus induces a partition $(X_0,X_1)$ of $X$.
Then
$\SubRiskRandAbb = \SubRiskRandAbb^{0} + \SubRiskRandAbb^{1}$,
where $\SubRiskRandAbb^i$ is the restriction of $\SubRiskRandAbb$ to $X_i$,
so that e.g.
$\SubRiskRandAbb^0_s = \indicator{s = 0} \cdot \mathbb{P}( \Ssf = 0 ) \cdot \SubRiskRandAbb_s$.
Now if $\RiskM$ were not convex it would not be subadditive and we would have
$\RiskM(\SubRiskRandAbb^{0} + \SubRiskRandAbb^{1})=\RiskM(\SubRiskRandAbb) > \RiskM(\SubRiskRandAbb^{0})+\RiskM(\SubRiskRandAbb^{1})$.
In other words, by splitting into subgroups we could automatically make our risk measure smaller, which is counter to what we wish to achieve.
Convexity is also desirable because, combined with F3, it preserves tractability of optimisation.

\emph{Positive Homogeneity} (F2) is desirable  but not
essential.
We would like our fairness measure to not vary in a manner that changes the optimal $f$ when $\ell$ varies in a manner that leaves the base problem invariant.
For example, if $\ell'=c \cdot \ell$ for some $c>0$ then obviously $\argmin_{f\in\Fscr} \Risk_{\ell}(f) = \argmin_{f\in\Fscr} \Risk_{\ell'}(f)$.
If $\Rcal$ is positively homogeneous, then
$\Rcal(\SubRiskRandAbb')= \Rcal(c \cdot \SubRiskRandAbb) = c \cdot \Rcal(\SubRiskRandAbb)$ and thus
$\argmin_{f\in\Fscr}\Rcal(\SubRiskRandAbb)=\argmin_{f\in\Fscr}\Rcal(\SubRiskRandAbb ')$.
Observe this last statement would remain true if $\Rcal$ was $k$-homogeneous, for any $k\ge 0$.
Whether a relaxation of (F2) adds any practical advantage is not yet understood.
Positive homogeneity does imply that the units of measurement of $\Rcal(\Zsf)$ are automatically the same as those for $\Zsf$.
The assumption of 1-homogeneity is also beneficial when analysing duality properties of risk measures; see~\citet[Section 6]{Rockafellar:2013aa}.

\emph{Monotonicity} (F3) is desirable because when combined with convexity (F1) it ensures that if $f \mapsto \Risk(f)$ is
convex, then so will be $f\mapsto\Rcal(\SubRiskRand{f})$; see \citep[Section 5]{Rockafellar:2013aa}, and part 3 of Theorem \ref{theorem:quadrangle-first-part}.
It is also intuitive that one's overall risk not increase if all subgroup risks are decreased.
We note that a similar monotonicity assumption, and its implications, were also employed in~\citet[Section 4]{Dwork:2018}.

\emph{Lower Semicontinuity} (F4) is a technical assumption that avoids problems with limits \citep{Rockafellar:2013aa}.

\emph{Translation invariance} (F5) is desirable because if we replace $\ell$ by $\ell+C$ we have not changed the unfairness at all, just the expected risk value.

\emph{Aversity} (F6) means that deviation from perfect fairness
(Definition~\ref{defn:perfect-utility}) is penalised; without this property we would not be capturing deviation from ideal fairness.

\emph{Law Invariance} (F7) means that $\Rcal$ only depends upon $\Zsf$ via its \emph{distribution} $\Pbb_\Zsf$  through an induced functional $\Fcal_\Pscr\colon\Pscr(S)\rightarrow\reals$.
For a fairness measure this would mean that the identity of each of the values of the sensitive feature do not matter, only the distribution of the risk variable $\SubRiskRandAbb(s)$ as a function of $s\in S$.
This is clearly a desirable attribute for a fairness measure.

\clearpage

\arxivVersion{
\section{Relationship to fairness methods based on inequality measures}
\label{app:inequality}
An approach to fair machine learning  similar to that based on  fairness risk measures
proposed in the body of the paper has been developed by
\citet{Speicher:2018,Heidari:2018,Heidari:2019}, based instead on the notion of an
\emph{inequality} measure. It turns out that this approach is closely
related to that proposed in the main body of the paper. In this technical appendix,
we explore the
relationship between the two methods by working out the relationship
between inequality measures and fairness risk measures. 

\emph{The appendix  is independent of the main paper in the sense that the paper can be
understood entirely without looking at this appendix.}  It is included for
completeness, and because we believe the results may be of independent
interest.

The literature
on inequality measures is quite formal (axiomatic) and consequently this
appendix is too. The conclusion we draw
(\S\ref{subsec:risk-and-inequality-consequences}) is essentially that the
requirements on a fairness risk measure are more stringent than those
usualy imposed upon
inequality measures: every fairness risk measure induces a ``nice''
inequality measure, but it is not the case that every nice inequality
measure induces a fairness risk measure. The additional constraints on fairness risk
measures (convexity, continuity and monotonicity) 
are exactly what one wants for the sake of solving fair machine
learning problems, since they guarantee computationally easy optimization problems.

\subsection{Inequality Measures}
We will show that the two approaches to solving fair ML problems, based
respectively on 
inequality measures and risk measures, are intimately related
by demonstrating that every fairness risk measure induces an inequality
measure compatible with the Lorenz ordering (defined below).  A weaker
converse result holds which demonstrates that the requirements on a
fairness risk measure are more stringent than those traditionally imposed
upon an inequality measure.  This result may be of interest in its own
right since the literatures on risk measures and inequality measures appear
to have not been explicitly connected before in this manner. There are a
four specific exceptions of which we are aware:
	\begin{itemize}
	\setlength{\itemsep=0pt}
		\item \citet{Bennett:2015aa} showed that some existing
			inequality measures can be derived from a model of
			choice under risk inspired by Harsanyi, especially
			his observation \citep{Harsanyi:1975aa} that Rawls'
			maximin principle corresponds ``to an expected
			utility evaluation based on a lottery that assigns
			probability 1 to the event that one assumes the
			identity of the worst-off individual in society.'' 
		\item \citet{Greselin:2015aa} unified a range of inequality
			measures and risk measures via a similar device
			--- by choosing what they call  ``societal
			references'' (e.g. a statistic of a population
			distribution such as a measure of centrality, or a
			tail measure) combined with distributions of
			personal gambles (that determine an individual's
			position on a population-based function).  They
			make explicit connections with aspects of coherent
			risk measures.
		\item \citet{Gajdos:2012aa} made a connection between
			random variables being ``less risky'' (meaning  derivable
			from comparator variables by adding zero mean
			noise), and measures of inequality.  However, they
			did not draw connections to the notion of \emph{risk
			measures}.  
		\item  It has also been shown that the
			mathematical notion underlying the Pigou-Dalton
			transfer principle for inequality measures (Schur
			convexity) is known to reduce to law invariance for
			coherent risk measures on atomless probability
			spaces \citep{Dana:2005aa,Grechuk:2012aa}, but
			these authors did not proceed to develop the more
			detailed connections we develop below.
	\end{itemize}
	None of these works have systematically related the
	\emph{axioms} for inequality measures to the axioms for risk as we
	do in this appendix, nor do they propose the formulaic correspondence
	between risk measures and inequality measures that we do (see
	(\ref{eq:ID-def})--(\ref{eq:RI-def})), 
	which
	however is implicit in the work of \citet{Kolm:1976aa,Kolm:1976ab}. 

Inequality measures\footnote{Sometimes called ``measures of inequality''
or ``inequality indices''.} have been investigated by ``welfare'' economists,
interested in such issues as the distribution of wealth or income. Their
perspective is intrinsically moral, rather than the more traditional view
of economics as aspirational physics
\citep{Mirowski:1989aa,Mirowski:1992aa}, and such an engagement with the
moral dimensions has been forcefully advocated recently
\citep{Shiller:2011aa}.  Formal measures of economic inequality were made
famous by Tony Atkinson's 1970 widely cited paper \citep{Atkinson:1970aa}.
Late in his career
\citet{Atkinson:2009aa}\footnote{
	The title of Atkinson's paper is ``economics as a moral science''.
	There are at least four other works with that exact title, whose
	pertinent common message we distill via the following four brief quotations:
	\begin{itemize}
	\setlength{\itemsep=0pt}
		\item ``Adam Smith, who has strong claim to being both the
			Adam and the Smith of systematic economics, was a
			professor of moral philosophy and it was at that
			forge that economics was made.
			Even when I was a student,
			economics was still part of the moral sciences
			tripos at Cambridge University. It can claim to be
			a moral science, therefore, from its origin, if
			for no other reason. Nevertheless, for many
			economists the very term `moral science' will seem
			like a contradiction'' \citep{Boulding:1969aa}.
		\item ``Economics is, and always has been, essentially a
			moral science whatever the protestations to the
			contrary by some of its practitioners''
			\citep{Cochran:1974aa}.
		\item ``[E]conomic science must merge with moral science''
			\citep[page 309]{Hodgson:2001aa}.
		\item ``Morality is not something we add to a model
			constructed without it; one built without morality
			as its integral component is an unsound structure.
			Restoring economics to its former habitat as a
			moral science, a science of practical knowledge,
			cannot be evaded''  \citep[page 9]{Rona:2017aa}.
	\end{itemize}
	Amartya Sen, in his  \emph{On Ethics and Economics},   argued that
	``economics has had two rather different origins, both related to
	politics, but related in rather different ways, concerned
	respectively with `ethics', on the one hand, and with what may be
	called `engineering', on the other'' \citep[pages
	2--3]{Sen:1987aa}.
	Sen uses ``ethics'' to mean choices of values and goals, and
	``engineering'' as mere means to achieve them. Ironically there has
	been a steady growth in soul searching in the discipline of
	engineering itself asking these same questions.  Most recently, the
	engineering field of machine learning is waking up to the same
	realisation; perhaps soon we shall see papers entitled
	\emph{Machine Learning as Moral Science}!
}
bemoaned the decline in explicit discussion of  welfare economics from its
heyday in the 1960s. Atkinson stressed the need for \emph{plurality} ---
that there is no single criteria that captures welfare: 
\begin{quote}
	Many of the ambiguities and disagreements [in economics] stem not from
	differences of view about how the economy works but about the criteria to
	be applied when making judgments. \ldots People can legitimately reach
	different conclusions because they apply different theories of justice.
	\citep[page 803]{Atkinson:2009aa}.
\end{quote}

The rest of the appendix is organised as follows. In Section 
\ref{subsec:axiomitising-inequality-measures} we introduce the main axioms
used in studying inequality measures; 
in \S\ref{subsec:relating-risk-and-inequality-axioms} we present a number
of apparently new results formally relating the axioms from the previous
subsection with those for fairness measures stated in the main body of the
paper; in \S\ref{subsec:risk-and-inequality-conclusions} we pull the
various lemmas together and state our main result; and finally in 
\S\ref{subsec:risk-and-inequality-consequences} we make some brief general
conclusions regarding the consequences for the use of either risk or
inequality measures to solve the problems of fair machine learning.

\subsection{Axiomatizing Inequality Measures}
\label{subsec:axiomitising-inequality-measures}
The idea of an inequality measure $\Ical$ is to measure the degree of
inequality of a population, say, in terms of incomes; that is some measure
of variability,  unevenness or non-uniformity. 
In contrast to risk measures, which can work
sensibly on continuous probability spaces, inequality measures are usually
only defined for populations of individuals of some finite size
$n\in\naturals$.  Let $x\in\reals^n$, and represent by $x_i$ the income of
the $i$th individual.   Without loss of generality we assume incomes are
nonnegative, and thus an \emph{inequality measure} is a function
$\Ical\colon\reals_{\ge 0}^n\rightarrow\reals_{\ge 0}$.  It is often
assumed that the ordering of the individuals is such that $i<j\Rightarrow
x_i\le x_j$. This assumption will be relaxed later by imposing a symmetry
condition on  $\Ical$.

Merely requiring $\Ical$ to be a function $\reals_{\ge
0}^n\rightarrow\reals_{\ge 0}$ is hardly illuminating or satisfactory. In
order to capture what is intuitively understood by ``inequality'' a range
of conditions are additionally imposed.  Depending upon which conditions
are used, this leads to a large variety of measures of inequality, some of
which are surveyed by \citet{Cowell:2000aa} and \citet{Sen:1997aa}.

In order to introduce these conditions on $\Ical$ and to 
relate inequality measures to fairness measures, we introduce
some notation and terminology due to \citet{Marshall:2011aa}.
Suppose $x\in\reals^n$. We write $x_{[i]}$  for  the
$i$th component of the \emph{decreasing} 
\emph{rearrangment}  of $x$ which satisfies $i<j\Rightarrow x_{[i]}\ge
x_{[j]}$.
For $x,y\in\reals^n$, we say $x$ is \emph{majorized} by $y$ on $A$  and write
$x\major y$ on $A$ if $x,y\in A$ and
\[
	\sum_{i=1}^k x_{[i]} \le \sum_{i=1}^k y_{[i]}, \forall k\in[n-1]\ \
	\ \mbox{and}\ \ \ \sum_{i=1}^n x_{[i]}=\sum_{i=1}^n y_{[i]}.
\]
Suppose $A\subset\reals^n$. A function $\phi\colon A\rightarrow\reals$ is
\emph{Schur-convex on $A$} if 
\[
	x\major y \mbox{\ on\ } A \ \ \Rightarrow  \ \ \phi(x)\le \phi(y) .
\]
If, furthermore, $\phi(x)<\phi(y)$ whenever $x\major y$ but $y$ is not a
permutation of $x$, then $\phi$ is \emph{strictly Schur-convex on $A$}. If
$A=\reals^n$ we simply say \emph{Schur-convex} or \emph{strictly
Schur-convex}.  Let $\Ccal(X)$ denote the set of continuous convex
functions on $X$.  We have \citep[page 14]{Marshall:2011aa} that 
\[
	x\major y
	\ \Leftrightarrow\  \sum_{i=1}^n \phi(x_i) \le \sum_{i=1}^n \phi(y_i) \ \ 
	\forall\phi\in\Ccal(\reals^n).
\]
Every convex symmetric function is Schur convex. A special case of this are
\emph{convex seperable} functions of the form $\phi(x)=\sum_{i=1}^n
g(x_i)$, where $g\colon\reals\rightarrow\reals$ is convex and continuous.
But not all Schur-convex functions are convex seperable.  The significance
of Schur-convexity for our current purpose is that the Pigou-Dalton
condition on an inequality measure $\Ical$ is equivalent to requiring that
$\Ical$ is strictly Schur-convex \citep[page 560]{Marshall:2011aa}.

A widely used notion in the theory of economic
inequality is the Lorenz curve. We associate with an income vector
$x\in\reals_{\ge 0}^n$ a probability distribution $\mu_x$ defined via 
\[
	\mu_x(\{x_i\})\coloneqq\textstyle\frac{1}{n} |\{j\in[n]\colon
	x_j=x_i\}|.
\]
Let $F_x$ denote the corresponding cumulative distribution function:
$F_x(t)\coloneqq\frac{1}{n} |{i\colon x_i\le t} |$ and $F_x^{-1}$ its quasi inverse
$F_x^{-1}(p) \coloneqq \inf \{x\colon F(x)\ge p\}$.  
The \emph{Lorenz curve} for $x$ is then defined as $\{(p,L_x(p)\colon
p\in[0,1]\}$; i.e.~the graph of the function
\[
	L_x(p) \coloneqq \frac{1}{m_x} \int_0^p F^{-1}(t) \dd t\ \ \ \ \ p\in[0,1],
\]
where $m_x= \int_{-\infty}^\infty t \dd F_x(t)$ is the corresponding mean.
Let $\reals^{\uparrow n}\coloneqq
\{(x_1,\ldots,x_n)\in\reals^n\colon x_1\le x_2\le\cdots \le x_n\}$. The 
Lorenz curve induces a partial order on $\reals^{\uparrow n}$ via
pointwise domination of the corresponding functions:
$x\preclorenzeq y \Leftrightarrow L_x \ge L_y$. This is called the
\emph{Lorenz ordering}.

We can interpret $x\in\reals^n$ as a map $x\colon[n]\rightarrow
\reals$ and thus for a permutation $\pi$, we write
$x\circ\pi=(x_{\pi(1)},\ldots,x_{\pi(n)})$. We denote the Kronecker product
by $\otimes$ and $1_r\in\reals^r$ the all ones vector of length $r$.
Let $x\in\reals^n$, $r\in\naturals$, and $x^{r}\coloneqq 1_r\otimes x$.  If
$S$ is a set and $J\in\naturals$, then $S_1,\ldots,S_J$ is called a 
\emph{partition} of $S$ 
if $S_i\cap S_j=\emptyset$ for $i,j\in [J]$, $i\ne j$, 
and $S=\bigcup_{j\in[J]} S_j$.

In order to make sense of the zoo of potential inequality measures, a range
of conditions (often called ``axioms'') are imposed 
\citep{Nunez-Velazquez:2006aa}.
We first state the axioms formally, and then
discuss them, providing the intuition and justification behind
them\footnote{We make no claim that the list of axioms presented here is
	complete. But they do appear to be the most significant and widely
	used ones}. In the inequality measure literature, often the domain
	is consistently taken to be $\reals^{\uparrow n}$ but since we will
	only consider measures that  satisfy symmetry, apart from axiom I1
	we presume $\Ical$ is defined on $\reals^n$.
\begin{description}
	\setlength{\itemsep=0pt} {}
	\item[I1\ \  Symmetry] For any permutation $\pi\colon[n]\rightarrow [n]$,
		$\Ical(x\circ\pi)=\Ical(x)$ for all
		$x\in\reals^{\uparrow n}$.
	\item[I2\ \  Scale Invariance]  For any $\lambda>0$, $\Ical(\lambda
		x)=\Ical(x)$ for all $x\in\reals^n$.
	\item[I3\ \  Pigou-Dalton Principle] $\Ical$ is strictly Schur-convex on
		$\reals^n$.
	\item[I4\ \  Dalton Population Principle] 
		$\Ical(x^{(r)})=\Ical(x)$ for all $x\in\reals^n$ and
		$r\in\naturals$.
	\item[I5\ \  Normalization] $\Ical(x)\ge 0$ for all
		$x\in\reals^{n}$ and $\Ical(x)=0\Leftrightarrow
		x=c 1_n$ for some $c\in\reals$.
	\item[I6\ \ Constant Addition] $\Ical(x+c 1_n) \le \Ical(x)$ for all
		$x\in\reals^{n}$ and $c>0$.
	\item[I7\ \  Lorenz Compatibility] $\Ical(x)\le \Ical(y) \Leftrightarrow
		x\preclorenzeq y$.
	\item[I8\ \  Additive Decomposition]  Let
		$J\in\naturals\setminus\{1\}$,  
		$S_1,\ldots,S_J$ be a
		partition of $[n]$, and for $j\in [J]$, let $n_j\coloneqq |S_j|$
		and
		$x^{(j)}\coloneqq (x_i)_{i\in S_j}$. An inequality
		measure $\Ical$ 
		satisfies \emph{additive decomposition} if
		\[
			(\forall x\in\reals^{\uparrow n})\ \ \  
			\Ical(x)=\sum_{j\in[J]} w_j \Ical(x^{(j)}) + B,
		\vspace*{-3mm}
		\]  
		where $w_j$ and 
		$B$ depend only on the 
		subgroup sizes $n_j$ and means
		$m_j\coloneqq\frac{1}{n_j}\sum_{i\in[n_j]} x^{(j)}_i$.
	\item[I9\ \  Subgroup Consistency]  Let $S_1,S_2$ be a partition of
		$[n]$.  An inequality measure $\Ical$ satisfies 
		\emph{subgroup consistency} if there exists an aggregation 
		function $\Phi$ increasing in its first two arguments, such that 
		\[
		     (\forall x\in\reals^{\uparrow n}) \  \ \ 
			\Ical(x) = \Phi\left(\Ical(x^{(1)}), \Ical(x^{(2)}); 
			m_1,m_2; n_1,n_2\right).
		\]
	\item[I10 Seperable] We say $\Ical$ is \emph{seperable} if 
		it can be written as $\Ical(x)=\sum_{i=1}^n g(x_i)$, where 
		$g\colon\reals_{\ge 0}\rightarrow\reals_{\ge 0}$. 
	\item[I11 Constant Sum Convexity] For $c>0$
		$T_c=\{x\in\reals^n\colon \sum_{i\in[n]} x_i = c\}$. Then
		$\Ical$ is \emph{constant sum convex} (resp.~\emph{constant
		sum strictly convex}) if for any $c>0$ it is
		convex (resp.~strictly convex) on $T_c$.
\end{description}
We make some comments regarding each of these axioms.
\begin{description}
	\setlength{\itemsep=0pt} {}
	\item[I1\ \  Symmetry] captures the notion that the identity of
		individuals should not change the perceived degree of
		inequality.
	\item[I2\ \  Scale invariance] is also known as 0-homogeneity or 
		homotheticity. It captures the idea that if all incomes
		(say) are multiplied by $\lambda>0$ then whilst the welfare
		and well-being of individuals would change, the inequality
		across the population would not.
	\item[I3\ \ Pigou-Dalton Principle] is usually described in terms of the
		``principle of transfers,'' whereby if income is transferred
		from a rich to a poor person in a ``mean-preserving''
		manner, then inequality can noot go up; see 
		\citep[page 6]{Marshall:2011aa} or \citep[page
		98]{Cowell:2000aa} for an exposition. For our purposes, the
		statement in terms of strict Schur convexity is more
		convenient.
	\item[I4\ \  Dalton Population Principle] is motivated by the idea
		that if a finite population is replicated $r$ times (for
		each individual with given wealth, for example, $r$ copies
		are created with the same wealth) then the inequality
		measure should not change.  This principle, as stated, hides a
		significant subtelty. Implicit in its statement is that we
		actually have a \emph{set} $\{\Ical^n\}_{n\in\naturals}$ 
		of inequality measures, where $\Ical^n\colon\reals_{\ge
		0}^n\rightarrow\reals_{\ge 0}$.  The relationship between
		$\Ical^n$ and $\Ical^m$ for $n\ne m$ obviously needs to be
		specified. Usually this is done by writing a functional
		form (i.e.~a symbolic formula) that holds for 
		all $n\in\naturals$. Of course if, for a fixed $n^*\in\naturals$,
		$\Ical^{n^*}$ is specified, then there is no unique way to
		induce the full set $\{\Ical^n\}_{n\in\naturals}$;
		confer \citep[page 165]{Chakravarty:1999aa}.  In some of the
		literature, this point is glossed over (e.g.
		\citep[page 205]{Nunez-Velazquez:2006aa}).
	\item[I5\ \  Normalization] guarantees the inequality measure is
		non-negative, and only equals zero (no inequality) when all
		$x_i$ ($i\in[n]$) are identical.
	\item[I6\ \ Constant Addition] captures the idea that a constant
		positive addition to everyone's income does not increase
		inequality.
	\item[I7\ \ Lorenz Compatibility] is of interest because of the
		historically wide use of the Lorenz curve and the induced
		Lorenz ordering; the axiom requires that the partial ordering
		induced by an inequality measure is the same as that
		induced by pointwise domination of the corresponding Lorenz
		curves.
	\item[I8\ \ Additive Decomposition] is motivated by the idea that
		the inequality of an entire population must be related to
		the inequalties of its constituent sub-populations. The
		condition is very strong and, in contrast say to I6, is
		stated as an \emph{equality} rather than as an \emph{inequality}. 
		\citet{Shorrocks:1980aa} has shown that when combined 
		with I2 and I4, the only admissible inequality measures are
		those in a single parameter generalised entropy family.
	\item[I9\ \ Subgroup consistency] introduced in
		\citep{Shorrocks:1984aa}, is a significant weakening of I8
		(I8 implies I9). Rather than insisting on an additive
		decomposition across subgroups, it is merely required that
		the inequalities of subgroups be aggregatable to recover
		the inequality of the whole population.  Surprisingly,
		there is a sort of converse whereby an inequality measure
		satisfying I9 can be transformed via a monotonic 
		transformation to one satisfying I8. Axiom I9 can be
		applied recursively to hold for partitions of arbitrary
		size.  The inequality in the definition of I9 excludes
		inequality measures with the property that inequality in
		every subgroup rises, but overall inequality falls.
		Properties of inequality measures from the perspective of
		subgroup decomposition are explored in depth by
		\citet{Deutsch:1999aa}.
	\item[I10\ \ Seperability] Is a special case of additive
		decomposition, and is an oft imposed condition on Social
		Welfare functions from which inequality measures can be
		derived \cite{Cowell:2000aa}.
	\item[I11\ \ Constant Sum Convexity]   Does not appear to be widely assumed in
		the inequality measure literature, with the exception of
		\citep[Section IX]{Kolm:1976ab} who provides 
		 an extensive justification of it and related notions of convexity.
		 Constant sum convexity is
	         arguably a better assumption than I8, I9, and I10
		 for three reasons.  First, it allows a neater
		 connection with risk measures. Second,  it is mathematically more
		 elegant. The third reason is
		 perhaps more interesting. Axioms I8 and I9 are stated as
		 \emph{equalities} and this allows the application of the
		 theory of functional equations in analysing the
		 implications.  Capturing the intent of I9 (one can not
		 make inequality look better by dividing into subgroups)
		 via an \emph{inequality} precludes analysis via the theory
		 of functional equations, but actually makes the arguments
		 easier via the theory of convexity. 
		 \citet[page 423]{Kolm:1976aa} also proposed subadditivity  as
		 an axiom. If $\Ical$ is assumed 1-homogeneous, this is
		 equivalent to convexity (recall [subadditivity and 
		 positive homogeneity] implies
		 [convexity and positive homogeneity]).
\end{description}

The motivation of scholars of inequality in adopting the axiomatic approach
varies. Many use it as a device to whittle down the large number of
possibilites to a few or even one single inequality measure. But the
arguments regarding the social desirability of the various axioms are not
so clear cut. \citet{Kolm:1976aa,Kolm:1976ab} in particular has argued at
length against a simple-minded acceptance of any of these axioms; confer
\citep{Cowell:2011aa}.  \citet{Kampelmann:2009aa} has observed that the
appeal of such analyses is sometimes that of a pursuit of objectivity in
making a choice that is seemingly ``opinion-free''.  More fundamentally, any notion
of \emph{inequality} is predicated on some notion of \emph{equality}, which
turns out to be more subtle than one might think.  Even in the realm of
pure mathematics it is a subtle concept, which \citet{Mazur:2008aa} has
argued is better construed as \emph{equivalence up to a canonical
isomorphism}.  One could even argue that the proposals of
\citet{Sen:1992aa} to take \emph{capabilities} into account is partially
grappling with this purely logical issue: taking income equality  as
``equality'' implies quotienting out everything else! (Confer
\cite{Loewenstein:2009aa} for a complementary argument against the notion 
that simple scalar measures can capture what is important.).

Fortunately, we can sidestep these weighty questions because our goal is
not to prescribe which are suitable for the social and economics purposes
to which they are typically applied, but rather to simply explore the
relationship with risk measures in order that we can sensibly compare fair
machine learning approaches built upon inequality measures with those built
upon risk measures. 

\subsection{Relating properties of risk measures and properties 
of inequality measures}
\label{subsec:relating-risk-and-inequality-axioms}
The theory of risk measures and the theory of inequality measures have been
developed independently. We will now show that there is a straight-forward
connection between them, which, up to some minor qualifications,  provides
a 1:1 correspondence between a fairness risk measure $\Rcal$ (or
equivalently a symmetric deviation measure $\Dcal$) and an inequality
measure $\Ical$.


In order to achieve this goal we first we need to deal with the different type signatures: $\Rcal$ and
$\Dcal$ take a random variable $\Xsf\colon S \rightarrow\reals_{\ge 0}$ as
an argument; inequality measures $\Ical$ take a vector 
$x\in\reals_{\ge 0}^{\uparrow n}$ or $\reals_{\ge 0}^n$ as an argument. 
We demand that $\Ical$, $\Rcal$ and $\Dcal$  all be
symmetric which means henceforth we can take the domain of $\Ical$ as
$\reals_{\ge 0}^n$.   Risk measures can be defined on an arbitrary sample
space $S$, but to construct the connection with inequality measures
we set $S=[n]$.  

Risk and deviation measures are defined on a space of random variables, and
there needs to be some base measure on the sample space $S$ for this to
make sense.  While as mentioned in the main body of the paper, the freedom
to choose different $\nu$ is a feature of the framework we propose, we will
restrict ourselves to $\nu=\nu_\unif^n$, the uniform probability measure on
$[n]$: for $B\subseteq [n]$, $\nu_\unif^n(B)=|B|/n$.   There are two
reasons for this.  First, it immediately guaramtees symmetry of the induced
inequality measures.  Second, one can extend the argument due to
\citet{Harsanyi:1955aa,Harsanyi:1958aa, Harsanyi:1975aa, Harsanyi:1978aa}
to justify the choice of a uniform distribution as being morally warranted
(when $S$ corresponds to sensitive features) a uniform distribution
corresponds to treating each value of the sensitive feature equally. In the
limit where each person who is represented by data corresponds to an
element of $S$, we precisely recover Harsanyi's setting and a uniform
distribution corresponds to the moral principle of treating people without
regard for their identity. 

A vector $x\in\reals_{\ge 0}^n$ can be identified with a a function
$[n]\rightarrow\reals_{\ge 0}$. We henceforth identify random variables
$\Xsf\colon[n]\rightarrow\reals_{\ge 0}$ with vectors $x\in\reals_{\ge
0}^n$ via the identity
\[
	x_i = \Xsf(i),\ \ \ i\in[n].
\]
Since our sample space $S$  is now assumed finite, there are no
measurability issues to concern us.  We thus write $\Ical(x)=\Ical(\Xsf)$
as synonyms; likewise we write $\Rcal(\Xsf)=\Rcal(x)$ and
$\Dcal(\Xsf)=\Dcal(x)$ as synonyms.

We know from the quadrangle theorem (Theorem~\ref{theorem:quadrangle-first-part})
that regular risk measures $\Rcal$ and regular deviation measures $\Dcal$
are in 1:1 correspondence via the relationship 
\begin{equation}
	\Rcal(\Xsf)= \Ebb(\Xsf)+\Dcal(\Xsf).
	\label{eq:R-via-D}
\end{equation}
Motivated by (\ref{eq:R-via-D}), we define an inequality measure $\Ical$ in terms of a
given $\Rcal$ or $\Dcal$, and conversely define a risk measure $\Rcal$ or
deviation measure $\Dcal$ in terms of a given inequality measure $\Ical$.
Given a deviation measure $\Dcal$, and a random variable $\Xsf\colon
S\rightarrow\reals_{\ge 0}$, let
\begin{equation}
	\label{eq:ID-def}
	\Ical_\Dcal(\Xsf) \coloneqq
	\begin{cases}\Dcal(\Xsf)/\Ebb(\Xsf), &\mbox{if\ } \Ebb(\Xsf)\ne 0\\
		0 & \mbox{if\ } \Ebb(\Xsf)=0
		 \end{cases}
\end{equation}
(Observe that since $\Xsf\ge 0$, we have that $\Ebb(\Xsf)=0 \Rightarrow
\Dcal(\Xsf)=0$ for any regular deviation measure, and thus the second case
above actually follows from the first by considering $\lim_{c\downarrow
0}\Dcal(\Xsf_c)/\Ebb(\Xsf_c)$, where $\Xsf_c:=c\Xsf$, and $\Xsf$ is an
arbitrary random variable such that $\Xsf\ge 0$ and $\Dcal(\Xsf)> 0$.)
Using (\ref{eq:R-via-D}), we can equivalently define
\begin{equation}
	\label{IR-def}
	\Ical_\Rcal(\Xsf) \coloneqq
	\frac{\Rcal(\Xsf)}{\Ebb(\Xsf)}-1,
\end{equation}
which is guaranteed to be well defined for any $\Rcal$ satisfying F6.
Conversely, given an inequality measure $\Ical$, we define
\begin{align}
	\Dcal_\Ical(\Xsf)& \coloneqq\Ebb(\Xsf)\Ical(\Xsf)\label{eq:DI-def} \\
	\Rcal_\Ical(\Xsf)&\coloneqq \Ebb(\Xsf)+\Ebb(\Xsf)\Ical(\Xsf) =
	\Ebb(\Xsf)[1+\Ical(\Xsf)].\label{eq:RI-def}
\end{align}
As an example, consider
$\Dcal(\Xsf)=\sigma(\Xsf)$ (the standard deviation), in which case we get
$\Ical_\sigma(X)=\sigma(\Xsf)/\Ebb(\Xsf)$, which is known as the
\emph{coefficient of variation}.  

We will now justify the above definitions by showing, under
suitable assumptions on $\Ical$, that $\Dcal_\Ical$ (resp.~$\Rcal_\Ical$)
is a deviation (resp.~risk) measure satisfying a subset of axioms
F1---F8; and conversely, given suitable assumptions on $\Dcal$
(resp.~$\Rcal$), that $\Ical_\Dcal$ (resp.~$\Ical_\Rcal$) is  an
inequality measure satisfying a subset of axioms I1--I11.

\subsubsection{I1 Symmetry}

The symmetry constraints on fairness measures and inequality measures are
easily related:
\begin{lemma}
	\label{lem:symmetry}
	Suppose $S=[n]$ and $\nu=\nu_\unif^n$.  
	\begin{enumerate}
		\item If $\Ical$ is symmetric (I1) then $\Rcal_\Ical$ is 
			law invariant (F7).
		\item If $\Rcal$ is law-invariant (F7) then $\Ical_\Rcal$ 
			is symmetric (I1).
	\end{enumerate}
\end{lemma}
For regular risk measures, law invariance of $\Rcal$ implies
law-invariance of $\Dcal$ so an analogous results holds with $\Rcal$
replaced by $\Dcal$ throughout.
\begin{proof}
	Let $\Pi_n$ be the set of all permutations $\pi\colon[n]\rightarrow
	[n]$.  Observe that for any $\pi\in\Pi_n$,
	$\nu_\unif^n\circ\pi^{-1}=\nu_\unif^n$.  Given that
	$\Rcal_\Ical(\Xsf)=\Ebb(\Xsf)[1+\Ical(\Xsf)]$, we have
	$\Rcal_\Ical$ is law-invariant (resp.~symmetric) if and only if
	$\Ical$ is law-invariant (resp.~symmetric) since $\Ebb$ is
	law-invariant (resp.~symmetric). 

	It convenient to rewrite the definitions of symmetry and
	law-invariance as follows.
	\begin{align}
		\Ical\mbox{\ is symmetric if\ }
			& \Ical(\Lsf)=\Ical(\tilde{\Lsf})\ \
			\forall\Lsf,\ \forall\tilde{\Lsf}\in\Lscr_{\mathrm{sym}}(\Lsf)
			\label{eq:rewritten-sym-def}\\
		\Ical\mbox{\ is law invariant if\ }
			& \Ical(\Lsf)=\Ical(\tilde{\Lsf})\ \
			\forall\Lsf,\ \forall\tilde{\Lsf}\in\Lscr_{\mathrm{li}}(\Lsf),
			\label{eq:rewritten-li-def}
	\end{align}
	where $ \Lscr_{\mathrm{sym}}(\Lsf)\coloneqq 
	\{\Lsf\circ\pi\colon\pi\in\Pi_n\}$ and
	$ \Lscr_{\mathrm{li}}(\Lsf)\coloneqq
		\{\Lsf\colon\mu_{\Lsf}=\mu_{\tilde{\Lsf}}\}$.
	Thus in order to prove the lemma it suffices to show that for any
	$\Lsf$, $\Lscr_{\mathrm{sym}}(\Lsf)=\Lscr_{\mathrm{li}}(\Lsf)$
	since then the conditions (\ref{eq:rewritten-sym-def}) and
	(\ref{eq:rewritten-li-def}) are identical.
	
	Suppose $\tilde{\Lsf}=\Lsf\circ\pi$ for some $\pi\in\Pi_n$. Then
	for any $A\in\Bscr(\reals_{\ge 0})$ (the Borel ssubsets of
	$\reals_{\ge 0}$),
	\begin{align*}
		\mu_{\tilde{\Lsf}}(A) &=(\nu_\unif^n\circ\tilde{\Lsf}^{-1})(A)\\
		&=(\nu_\unif^n\circ\pi^{-1}\circ\Lsf^{-1})(A)\\
		&=(\nu_\unif^n\circ\Lsf^{-1})(A)\\
		&=\mu_{\Lsf}(A).
	\end{align*}
	Thus
	$\Lscr_{\mathrm{sym}}(\Lsf)\subseteq\Lscr_{\mathrm{li}}(\Lsf)$.

	Suppose instead that
	$\Lsf,\tilde{\Lsf}\colon[n]\rightarrow\reals_{\ge 0}$ are  such
	that $\mu_{\Lsf}=\mu_{\tilde{\Lsf}}$. Then the image of $S$ under
	$\Lsf$ and $\tilde{\Lsf}$ is identical: $\Lsf(S)=\tilde{\Lsf}(S)$.
	Let $\{t_1,\ldots,t_k\}=\Lsf(S)$ where $k\le n$. For $i\in[k]$,
	$\mu_{\Lsf}(\{t_i\})=j_i/n$, where $j_i$ is positive integer and
	$\sum_{i\in[k]} j_i=n$. Each elementary probability mass $1/n$
	corresponds under $\Lsf$ (resp.~$\tilde{\Lsf}$) to a particular
	$s_r\in S$ (resp.~$s_{\tilde{r}}\in S$), $r,\tilde{r}\in[n]$.  The
	only freedom in choosing $\tilde{\Lsf}$ and $\Lsf$ is in the
	indexing of elements of $S$.  Thus one can write
	$\tilde{r}=\pi^{-1}(r)$ for some permutation $\pi$ and thus
	$\tilde{\Lsf}=\Lsf\circ\pi$.  Consequently
	$\Lscr_{\mathrm{sym}}(\Lsf)\supseteq\Lscr_{\mathrm{li}}(\Lsf)$
	which completes the proof.
\end{proof}

\subsubsection{I2  Scale Invariance}
The form of scaling behaviour of inequality measures and risk measures is
also easily related:
\begin{lemma}
	\label{lemma:scale-invariance-relationship}
	If $\Dcal$ (resp.~$\Rcal$) is positively homogeneous
	(i.e.~satisfies F2) then $\Ical_\Dcal$ (resp.~$\Ical_\Rcal$) is
	0-homogeneous (I2);  conversely, if $\Ical$ satisfies I2 then
	$\Dcal_\Ical$ (resp.~$\Rcal_\Ical)$ satisfies F2.
\end{lemma}
\begin{proof}
	Since $\Ebb$ is positively homogeneous and the ratio of two
	positively homogeneous functions is 0-homogeneous, we obtain the
	first part. The second part follows since the product of a
	1-homogeneous function with a 0-homogeneous function is
	1-homogeneous.
\end{proof}
The fact that if $\Ical$ is 0-homogeneous then $\Xsf\mapsto
\Ebb(\Xsf)\Ical(\Xsf)$ is 1-homogeneous was observed by \citet[page
423]{Kolm:1976aa}.

\subsubsection{I3 Pigou-Dalton (Schur-Convexity)}
We need the following straightforward lemma.
\begin{lemma}
	\label{lemma:schur-convex-normalisation}
	Suppose $\phi\colon\reals^n\rightarrow\reals$. Let
	$\bar{x}\coloneqq\frac{1}{n}\sum_{i=1}^n x_i$ and define
	$\tilde{\phi}\colon x\mapsto \phi(x)/\bar{x}$.  Then $\tilde{\phi}$
	is Schur-convex (resp.~strictly Schur-convex) if and only if 
	$\phi$ is Schur-convex (resp.~strictly Schur-convex).
\end{lemma}
\begin{proof}
	Suppose $x,y\in\reals^n$ and $x\major y$ which implies
	$\bar{x}=\bar{y}=:m$.  Suppose $m >0$ (we justify that this
	does not lose generality below). 
	Let $\tilde{x}\coloneqq x/m$ and $\tilde{y}=y/m$. Observe that
	$x \major y \Leftrightarrow \tilde{x}\major\tilde{y}$. 
	We thus have
	\begin{align*}
	\phi \ \mbox{is Schur-convex}
	\Leftrightarrow\ & [x\major y  \Leftrightarrow \phi(x) \le \phi(y)]\\
	\Leftrightarrow\ & [\tilde{x}\major \tilde{y}  \Leftrightarrow
		\phi(\tilde{x}) \le \phi(\tilde{y})]\\
	\Leftrightarrow\ & [{x}\major {y}  \Leftrightarrow
		\tilde{\phi}({x}) \le \tilde{\phi}({y})]\\
	\Leftrightarrow\ & \tilde{\phi} \ \mbox{is Schur-convex.}
	\end{align*}
Since [$x$ is not a permutation of $y$] $\Leftrightarrow$ [$\tilde{x}$ is
not a permutation of $\tilde{y}$] we similarly can conclude that strict
Schur-convexity of $\phi$ is equivalent to strict Schur-convexity of
$\tilde{\phi}$.
\end{proof}
\begin{lemma} 
	\label{lem:induced-schur-convexity}
	Suppose $\Dcal$ is convex (F1), then $\Ical_\Dcal$ is strictly
	Schur-convex (I3).
\end{lemma}
\begin{proof}
	Combine Lemma~\ref{lemma:schur-convex-normalisation} with the fact
	that every convex symmetric function is Schur-convex
	\citep{Marshall:2011aa}.
\end{proof}
A similar result has been observed in a particular case that complements
the situation of interest to us: a probability space $S$ is called
\emph{atomless} is there exists a random variable on $S$ with a continuous
cumulative distribution function.  \citet{Dana:2005aa} showed (confer
\cite{Grechuk:2012aa}) that on an atomless probability space, every law
invariant risk functional $\Lcal^p(S)\rightarrow \bar{\reals}$ is
Schur-convex.   A converse to the above lemma is impossible because there
are Schur-convex functions that are not convex \citep{Marshall:2011aa}.

\subsubsection{I4   Dalton Population Principle} 
\begin{lemma}
	\label{lemma:dpp}
	If $\Rcal$ (resp. $\Dcal$) is law invariant (F7), then  
	$\Ical_\Rcal$ satisfies I4.
\end{lemma}
\begin{proof}
	It follows from the definition of $\Ical_\Rcal$ that $\Ical_\Rcal$
	satisfies I4 if and only if $\Rcal$ does (since expectation
	trivially does not change under replication). For $r\in\naturals$,
	given a random variable $\Lsf\colon[n]\rightarrow\reals_{\ge 0}$
	define the $r$-replicated random variable
	$\Lsf_{(r)}\colon[rn]\rightarrow\reals_{\ge 0}$ via $
	\Lsf_{(r)}(i)\coloneqq \Lsf(\psi_r(i))$, where
	$\psi_r\colon\naturals\rightarrow\naturals$ is given by
	$\psi_r(i)\coloneqq\lceil i/r\rceil$. The inverse
	$\psi_r^{-1}\colon\naturals\rightrightarrows\naturals$ is given by
	$\psi_r^{-1}(j)=\{s\in\naturals\colon \lceil
	s/r\rceil=j\}=\{rj,rj+1,\ldots,rj+(r-1)\}$.  Thus the probability
	law of $\Lsf_{(r)}$ is given by (for $A\in\Bscr(\reals_{\ge 0})$)
	\begin{align*}
		\mu_{\Lsf_{(r)}}(A)&= (\nu_\unif^{nr}\circ\Lsf_{(r)}^{-1})(A)\\
		&=(\nu_\unif^{nr}\circ\psi_r^{-1}\circ\Lsf^{-1})(A)\\
		&=(\nu_\unif^{n}\circ\Lsf^{-1})(A)\\
		&=\mu_{\Lsf}(A),
	\end{align*}
	because $\nu_\unif^{nr}\circ\psi_r^{-1}=\nu_\unif^n$.  Since $\Ical$
	is law invariant, 
	$\Ical_\Rcal(\Lsf_{(r)})=\Ical_{\Rcal}(\Lsf)$ and consequently 
	$\Ical_\Rcal$ satisfies the Dalton Population Principle I4.
\end{proof}
The Dalton Population Principle does not, on its own, imply law invariance.
It can't, because if we start with a population of size $n$, it only makes
assertions about populations of size $nr$ for $r\in\naturals$. But given a
distribution $\mu_\Lsf$, there are many more sample spaces (not necessarily
of cardinality $nr$) on which one can define random variables
$\tilde{\Lsf}$ that can give rise to the same distribution.   In
particular, if one does not impose symmetry, one can construct examples of
$\Ical$ that are non-symmetric which satisfy the Dalton population principle
but are not law-invariant.

The condition of $\Ical$ being law-invariant also has the subtlety of
domain of definition that the Dalton population principle has, but this can
be cleanly swept under the carpet as follows. Let $(S,\Sscr,\nu)$ be a
probability space and let
$\Mscr_S\coloneqq\{\Lsf\colon(S,\Sscr)\rightarrow(\reals_{\ge
0},\Bscr(\reals_{\ge 0}))\}$ denote the set of random variables defined on
$(S,\Sscr,\nu)$, measureable with respect to the Borel sigma-algebra on
$\reals_{\ge 0}$. For a \emph{given} $S$ (with associated $\Sscr$ and
$\nu$), one can talk precisely of $\Ical\colon\Mscr_S\rightarrow\reals_{\ge
0}$. However, one can \emph{not} talk sensibly of ``$\Ical\colon \bigcup_S
\Mscr_S\rightarrow\reals_{\ge 0}$'' where the putative union is over ``all
sample spaces'' (i.e. over all possible sets, with all the difficulties
such a notion implies).  However, there is a simple fix that avoids such a
definitional conundrum: the whole point of the notion of law invariance is
that $\Ical$ depends upon $\Lsf\colon S\rightarrow\reals_{\ge 0}$
\emph{only} in terms of the probability distribution
$\mu_{\Lsf}(A)=(\nu\circ\Lsf^{-1})(A)$. That is, one can write
$\Ical(\Lsf)=\bar{\Ical}(\mu_\Lsf)$ for some function $\bar{\Ical}$.
Whilst to \emph{compute} $\mu_\Lsf$ one needs to specify a sample space $S$
(and base measure $\nu$) the distribution itself is simply a function
$\mu_{\Lsf}\colon\Bscr(\reals_{\ge 0})\rightarrow [0,1]$ --- and the sample
space becomes invisible.  We could thus construe $\Ical$, $\Rcal$ and $\Dcal$
as having the type signature $\Ical,\Rcal,\Dcal\colon \Delta(\reals_{\ge
0})\rightarrow\reals_{\ge 0}$, where $\Delta(\reals_{\ge 0})$ denotes the
set of probability distributions on $\reals_{\ge 0}$.  (Confer the
arguments of \citet[Chapter 1]{Le-Cam:1986aa} regarding avoiding explicit
use of the sample space.)

\subsubsection{I5 Normalization} 
We restate I5 in the language of random variables as: $\Ical(\Xsf)\ge 0$
for all $\Xsf\colon[n]\rightarrow\reals_{\ge 0}$ and $\Ical(\Xsf)=0$ if and
only if $\Xsf=\Csf$ where $\Csf$ is a constant random variable, $\Csf(i)=C$
for all $i\in[n]$ for some $C\in\reals_{\ge 0}$.
\begin{lemma}
	\label{lemmma:normalization}
		 Suppose $\Ical$ satisfies I5, then $\Rcal_\Ical$ satisfies
		 F6 and F8.  Conversely, Suppose $\Rcal$ satisfies F6 and
		 F8. Then $\Ical_\Rcal$ satisfies I5.
\end{lemma}
\begin{proof}
	$\Ical$ satisfying I5 implies $[\Rcal_\Ical(\Xsf)=
	\Ebb(\Xsf)(1+\Ical(\Xsf))\ge 0 \ \
	\forall\Xsf\colon[n]\rightarrow\reals_{\ge 0} ] $ and $
	[\Rcal_\Ical(\Xsf)=\Ebb(\Xsf)\mbox{\ \ if and only if\ \ }
	\Xsf=\Csf].  $ Thus $\Rcal_\Ical$ satisfies F6. Furthermore we have
	$\Rcal(\Csf)=\Ebb(\Csf)[1+\Ical(\Csf)]=C$ and so $\Rcal_\Ical$
	satisfies F8.

	Conversely suppose $\Rcal$ satisfies F6 and F8 and
	$\Ical_\Rcal(\Xsf)=\frac{\Rcal(\Xsf)}{\Ebb(\Xsf)}-1$. Suppose
	$\Xsf$ is constant ($\Xsf=\Csf$), then F8 implies $\Rcal(\Csf)=C$.
	But $\Ebb(\Csf)=C$ also  and thus $\Ical_\Rcal(\Xsf)=0$.
	Alternatively, if $\Xsf$ is non-constant, then by F6, we have
	$\Ical_\Rcal(\Xsf)> 0$.  Thus $\Ical_\Rcal$ satisfies I5.
\end{proof}
Since we know (main body of paper) that F5 and F6 imply F8, we also have
the second part of above lemma holding where the assumption is simply F5.
\subsubsection{I6 Constant Addition} 
\begin{lemma}
	\label{lemma:constant-addition}
	Suppose $\Rcal$ satisfies F5 and [F6 or (F1 and F2) or
	(Subadditivity and F2)], then $\Ical_\Rcal$ satisfies I6.
	Conversely, suppose $\Ical$ satisfies I5, then $\Rcal_\Ical$
	satisfies F6.
\end{lemma}
\begin{proof}
	Translating I6 to the language of random variables we require $
	\Ical_\Rcal(\Xsf+\Csf) \le \Ical_\Rcal(\Xsf)$ for all
	$\Xsf\colon[n]\rightarrow\reals_{\ge 0}$ and all constant random
	variables $\Csf(i)=C$ for all $i\in[n]$ for some constant $C>0$.
	If $\Xsf(i)=0$ for $i\in[n]$, then F6 implies
	$\Rcal(\Xsf)=\Ebb(\Xsf)$ and thus
	$\Ical_\Rcal(\Xsf+\Csf)=\Ical_\Rcal(\Csf)=0=\Ical_\Rcal(\Xsf)$ and
	I6 is satisfied. Alternatively, if $\Xsf>0$ then $\Ebb(\Xsf)>0$ and we have 
	\begin{align*}
		\mathrm F6 \ \Rightarrow\ \ & \Ebb(\Xsf) \le \Rcal(\Xsf) \ \ \
	&\forall\Xsf>0\ \ \ \ \ \ \ \ \ \ \ \ \ \ \ \ \ \ \ \ \  \ \ \  \\
		\Rightarrow\ \ & C(\Ebb(\Xsf)-\Rcal(\Xsf))\ge0 \ \ \
			&\forall\Xsf>0,\ \forall C\in(0,\infty)\\
		\Rightarrow\ \ & \frac{C(\Ebb(\Xsf)
		      -\Rcal(\Xsf))}{\Ebb(\Xsf)(\Ebb(\Xsf) +C)} \le 0
		      \ \ \ &\forall\Xsf>0,\ \forall C\in(0,\infty)\\
	      \Rightarrow\ \ &
		      \frac{\Rcal(\Xsf)\Ebb(\Xsf)+C\Ebb(\Xsf)
			      -\Rcal(\Xsf)\Ebb(\Xsf) -
		      \Rcal(\Xsf)C}{\Ebb(\Xsf)(\Ebb(\Xsf)+C)} \le 0
		      \ \ \ &\forall\Xsf>0,\ \forall C\in(0,\infty)\\
	      \Rightarrow\ \ & \frac{\Rcal(\Xsf)+C}{\Ebb(\Xsf)+C} 
		      - \frac{\Rcal(\Xsf)}{\Ebb(\Xsf)} \le 0
		      \ \ \ &\forall\Xsf>0,\ \forall C\in(0,\infty)\\
	      \Rightarrow\ \ &  \frac{\Rcal(\Xsf+\Csf)}{\Ebb(\Xsf+\Csf)} -1 
		      \le \frac{\Rcal(\Xsf)}{\Ebb(\Xsf)} -1
		      \ \ \ &\forall\Xsf>0,\ \forall C\in(0,\infty),\\
	      \intertext{where pulling the $C$ into the argument of $\Rcal$
		      is justified by  F6 or (F1 \& F2) or
		      (Subadditivity \& F2)}
	      \Rightarrow\ \ & \Ical_\Rcal(\Xsf+\Csf) \le \Ical_\Rcal(\Xsf)
		      \ \ \ &\forall\Xsf>0,\ \forall C\in(0,\infty)\\
	      \Rightarrow \ \ &\mbox{I6}.
	\end{align*}
	Conversely, given $\Ical$, consider 
	$\Rcal_\Ical(\Xsf)=\Ebb(\Xsf)(1+\Ical(\Xsf))$. If $\Xsf=\Csf$ is
	constant then
	$\Rcal_\Ical(\Csf)=\Ebb(\Csf)(1+\Ical(\Csf))=\Ebb(\Csf)=C$, which
	proves one part of F8.
	If $\Xsf$ is not constant, then by I5, $\Ical(\Xsf)>0$ and thus
	$\Rcal_\Ical(\Xsf)> \Ebb(\Xsf)$ which
	proves the other part of F8.
\end{proof}

\subsubsection{I7 Lorenz Compatibility} 
\citet{Foster:1985aa} characterised Lorenz compatibility of inequality
measures via the following:
\begin{lemma}
	\label{lemma:lorenz-compatibility-characterisation}
	An inequality measure is Lorenz
		compatible (I7)  if and only if it satisfies I1, I2, I3 \& I4.
\end{lemma}
Relations between I7 and  F1,\ldots,F9 thus follow from 
Lemma 
(\ref{lemma:lorenz-compatibility-characterisation}) and the earlier lemmas.

\subsubsection{I8, I9, I10 and I11 (Decomposability)} 
We consider I8--I11 together. It is now convenient to write  inequality
measures (and deviation and risk measures) in terms of vectors
$x\in\reals_{\ge 0}^n$.  We write $\Ebb(x)=\|x\|_1/n$, where
$\|x\|_1=\sum_{i=1}^n x_i$ for $x\in\reals_{\ge 0}^n$.

We need the following lemma 
\citep[page 92, C.1.a]{Marshall:2011aa} applied to the interval
$(0,\infty)$:
\begin{lemma}
	\label{lemma:seperability-schur-convexity}
	Suppose $\Ical$ is seperable (I10). Then $\Ical$ is strictly Schur-convex if
	and only if $g$ is strictly convex.
\end{lemma}
A consequence of this lemma is that if $\Ical$ is seperable and satisfies
I3, then $\Ical$ is strictly convex.
We consider deviation measures induced from an inequality measure:
$\Dcal_\Ical(x)\coloneqq\Ebb(x)\Ical(x)$.  

\begin{lemma}
	\label{lemmma:decomposability}
	Suppose $\Ical$ satisfies [I2 and I3 and I10] or 
	[I2 and I11 (strict)].
	Then $\Dcal_\Ical$ (resp.~$\Rcal_\Ical$) is strictly convex
	and positively homogeneous (F1 and F2).
\end{lemma}
\begin{proof}
	For the first case, by Lemma
	\ref{lemma:seperability-schur-convexity}, $\Ical$ is strictly convex.  
	The second case simply assumes constant sum strict convexity of $\Ical$, so we can
	henceforth presume it.
	For $m>0$ let  $A_m\coloneqq\{x\in\reals_{\ge
	0}^n\colon\Ebb(x)=m\}$.  We first show that $\Dcal_\Ical$ is
	strictly convex
	on $A_m$ for all $m>0$. Fix $m>0$ and pick $x,y\in A_m$.  Then for
	any $\lambda\in(0,1)$,
	\begin{align*}
		\Dcal_\Ical(\lambda x+(1-\lambda)y) 
		&=\Ebb(\lambda x+(1-\lambda y)) \Ical(\lambda x+(1-\lambda)) y)\\
		&= m \Ical(\lambda x+(1-\lambda)y)\\
		& < m(\lambda\Ical(x)+(1-\lambda)\Ical(y))\\
		&= \lambda m \Ical(x)+(1-\lambda) m \Ical(y)\\
		&= \lambda \Dcal_\Ical(x) +(1-\lambda)\Dcal_\Ical(y),
	\end{align*}
	and thus $\Dcal_\Ical$ is strictly convex on $A_m$ for all $m>0$.
	Lemma \ref{lemma:scale-invariance-relationship} implies that
	$\Dcal_\Ical$ is 1-homogeneous on $\reals_{\ge 0}^n$.  We now show that
	these two facts imply $\Dcal_\Ical$ is strictly convex on 
	$\reals_{\ge 0}^n$.

	Pick $x,y\in\reals_{\ge 0}^n$ and let $m_x\coloneqq\Ebb(x)$ and
	$m_y\coloneqq\Ebb(y)$.  Observe that
	$\Ebb(\lambda x+(1-\lambda)y)= \lambda m_x +(1-\lambda) m_y$.
	Furthermore, since $\Dcal_\Ical$ is 1-homogeneous, for all
	$x\in\reals_{\ge 0}^n$, $\Dcal_\Ical(x) = \|x\|_1 \left(\frac{1}{\|x\|_1}
	\Dcal_\Ical(x)\right)=\|x\|_1 \Dcal_\Ical(x/\|x\|_1)$.  Thus for all
	$\lambda\in(0,1)$
	\begin{align*}
		\Dcal_\Ical(\lambda x+(1-\lambda) y) & = \|\lambda x+(1-\lambda)
		y\|_1 \Dcal_\Ical\left(\frac{\lambda x+(1-\lambda)y}{\|\lambda
		x+(1-\lambda)y\|_1}\right)\\
		&=  (\lambda n m_x +(1-\lambda)n m_y) \Dcal_\Ical(\lambda \bar{x}
		+(1-\lambda)\bar{y}),\\
		\intertext{where $\bar{x}\coloneqq x/(\lambda n m_x
			+(1-\lambda) n m_y)$ and $\bar{y}\coloneqq y/(\lambda 
			n m_x +(1-\lambda) n m_y)$ and $\bar{x},\bar{y}\in A_n$.
			Since $\Dcal_\Ical$ is strictly convex on $A_n$ we have}
	\Dcal_\Ical(\lambda x+(1-\lambda) y)  & < (\lambda n m_x +(1-\lambda)n
	m_y)\Dcal_\Ical(\bar{x}) + (1-\lambda) (\lambda n m_x +(1-\lambda) n m_y)
	\Dcal_\Ical(\bar{y})\\
	&= \lambda \Dcal_\Ical(x) + (1-\lambda) \Dcal_\Ical(y),
	\end{align*}
	and thus $\Dcal_\Ical$ is strictly convex.
	Since $\Rcal_\Ical(x)=\Ebb(x)+\Dcal_\Ical(x)$, strict convexity and
	positive homogeneity of $\Rcal_\Ical$ follows from the fact that
	these properties are preserved under summation with the convex and
	positively homogeneous function $x\mapsto\Ebb(x)$.
\end{proof}
If $\Ical$ is not seperable, then I3 does not guarantee strict
convexity (or even convexity) of $\Ical$ since not every strictly
Schur-convex function is strictly convex in which case we can not guarantee
the strict convexity (or even convexity) of $\Dcal_\Ical$.

\cite{Kolm:1976ab} has argued that one could equally normalise inequality
measures via 1-homogeneity, in which case the bridge to risk and deviation
measures is even simpler --- such an 1-homogeneous inequality measure \emph{is} a
deviation measure.


\subsection{Relating inequality measures and risk measures}
\label{subsec:risk-and-inequality-conclusions}
Let $\frm$ denote the set of fairness risk measures (i.e. $\Rcal$
satisfying F1--F7) and let $\stim$ denote the set of standardised inequality
measures (i.e. satisfying Lorenz compatibility I7, normalisation I5 and
constant addition I6).  Let
$\Rsim \coloneqq\{\Rcal_\Ical\colon\Ical\in\stim\}$.
\begin{theorem}
	Suppose $S=[n]$ and $\nu=\nu_\unif^n$ and that $\Rcal_\Ical$ and
	$\Ical_\Rcal$ are defined by (\ref{eq:ID-def})--(\ref{eq:RI-def}).
	\begin{enumerate}
		\item If $\Rcal\in\frm$ 
			then $\Ical_\Rcal\in\stim$.
		\item  $\Ical\in\stim$ then $\Rcal_\Ical$ satisfies
			F2,  F6, F7 and F8.  If furthermore $\Ical$ satisfies I10,
			then $\Rcal_\Ical$ also satisfies F1.
	\end{enumerate}
	Thus $\frm\subset\Rsim$.
\end{theorem}
\begin{proof}
	Part 1 follows from lemmas  
	\ref{lem:symmetry},
	\ref{lemma:scale-invariance-relationship},
	\ref{lem:induced-schur-convexity},
	\ref{lemma:dpp},
	\ref{lemmma:normalization} and
	\ref{lemma:constant-addition}.
	Part 2 follows from lemmas 
	\ref{lem:symmetry},
	\ref{lemma:scale-invariance-relationship},
	\ref{lemmma:normalization},
	\ref{lemma:constant-addition} and
	\ref{lemmma:decomposability}.
\end{proof}
The theorem suggests that fairness risk measures are a stronger notion than
inequality measures. We now give some intuition as to why it is not
plausible that the second part of the theorem can be strengthened, and that
fairness risk measures are indeed a stronger (more restrictive) notion.

A crucial property of fairness risk measures is monotonicity (F3). Observe
that if F3 is satisfied, not only do we have that (translating to vector
notation for now) for $x,y\in\reals_{\ge 0}^n$, $x\le y \Rightarrow
\Rcal(x)\le \Rcal(y)$ but also $x\le y \Rightarrow \Ebb(x)\le\Ebb(y)$.
Since $\Ical_\Rcal(x)=\frac{\Rcal(x)}{\Ebb(x)}-1$,  we can conclude nothing
about $\Ical(x+y)$ on the basis of these assumptions.
Furthermore, to consider the converse direction, the
crucial property of strict Schur-Convexity of $\Ical$ is equivalent to the
requirement that $x\major y \Leftrightarrow \Ical(x)\le\Ical(y)$ for all
such $\Ical$. But the relationship of majorisation conflicts with
the pointwise domination required for monotonicity since
if $x\le y$ and $x\major y$ then $x=y$ \citep[page 13]{Marshall:2011aa}.
Thus the property of pointwise inequality $x<y$ is ``invisible'' to inequality measures.
This failure matters for our motivating purpose --- to find learned
hypotheses that performs well in the traditional sense (average losses are
small) \emph{and} satisfies some notion of equity or fairness. While inequality
measures can judge the unfairness, if one combines them with the average loss
in a dimensionally sensible way (such as our 1-homogeneous proposal $\Rcal_\Ical$),
the resulting risk measure is not guaranteed to satisfy the highly desirable property of
monotonicity.

Although most specific proposed inequality measures are continuous, continuity does
not feature in the list of axioms typically applied to inequality measures.
And thus we can not guarantee that $\Rcal_\Ical$ satsifies F4. Of course
one could trivially impose it upon $\Ical$ and the same continuity would
immediately hold for $\Rcal_\Ical$.

Finally there is an intrinsic difficulty in deriving an \emph{equality}
constraint (F5) on $\Rcal_\Ical$ from an \emph{inequality} constraint on
$\Ical$ such as I6.  

Thus in general we can not guarantee that
$\Rcal_\Ical$ satisfies F3, F4 and  F5 and so $\Rcal_\Ical$ is not guaranteed to
be a fairness risk measure.

\subsection{Consequences for Fair Machine Learning and Beyond}
\label{subsec:risk-and-inequality-consequences}

Our conclusion from the above analysis is that for the purposes of learning
hypotheses from empirical data that perform well in terms of expected loss
and allow the control of fairness, it is better to use
the slightly more restricted class of
fairness risk measures $\frm$ rather than attempting to work with the larger
class $\Rsim$ which lacks many of the desirable properties of a
fairness risk measure (convexity, monotonicity, and continuity).   
Since $\Rcal\in\frm$ are also attractive from a computational standpoint,
fairness risk measures seem to be preferrable to inequality measures as a
basis for fair machine learning.

%

\clearpage
}
{}

\section{Additional experiments}
\label{app:experiments}

We present some experimental results supplementing those in the body.

\subsection{Results with real-valued sensitive feature}

\begin{figure*}[!h]
	\centering
	\includegraphics[scale=0.28]{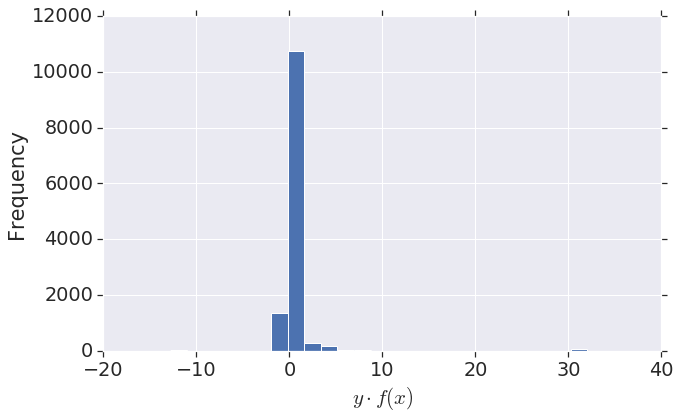}
	\quad
	\includegraphics[scale=0.28]{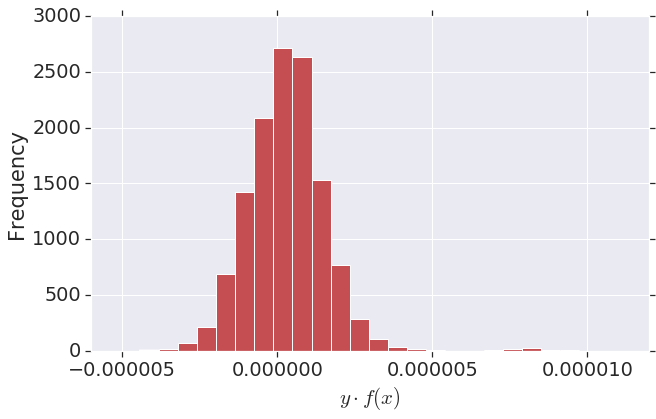}

	\caption{Results on {\tt adult} dataset with {\tt fnlwgt} as the continuous sensitive feature.
	The left and right panel are for $\alpha = 0.1$ and $\alpha = 0.9$ respectively.
	Since essentially each individual is a member of a singleton subgroup, the latter is seen to encourage commensurate model predictions, and thus margin scores across all instances.
	}
	\label{fig:adult-continuous}
\end{figure*}

We illustrate the viability of using the model with a real-valued sensitive feature $S$.
We consider the {\tt adult} dataset, but this time with {\tt fnlwgt} (an estimate of how representative an individual is) as the sensitive feature.
Following~\ref{eqn:cvar-continuous-empirical}, essentially all instances are placed into separate subgroups in forming the CVaR objective.
Figure \ref{fig:adult-continuous} compares the histogram of margin scores $\{ y_i \cdot f( x_i ) \}_{i = 1}^m$ for $\alpha = 0.1$ and $\alpha = 0.9$.
We see that, as expected, setting $\alpha = 0.9$ encourages all scores to be roughly commensurate.

\subsection{Additional results on {\tt synth} and {\tt adult}}

\begin{figure*}[!p]
	\centering
	\includegraphics[scale=0.28]{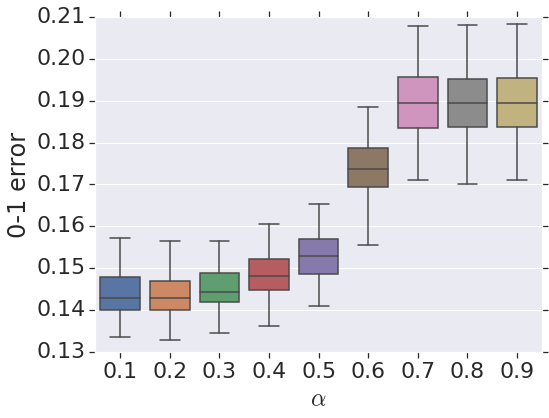}
	\quad
	\includegraphics[scale=0.28]{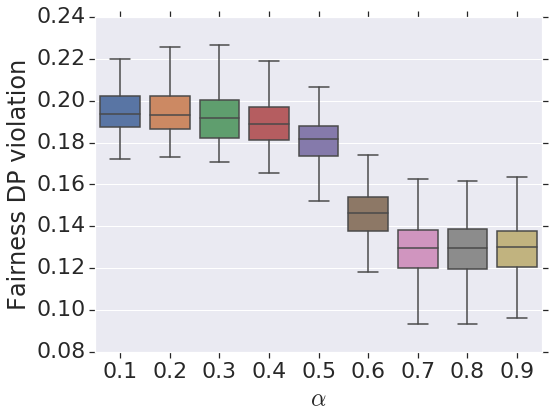}	

	\includegraphics[scale=0.27]{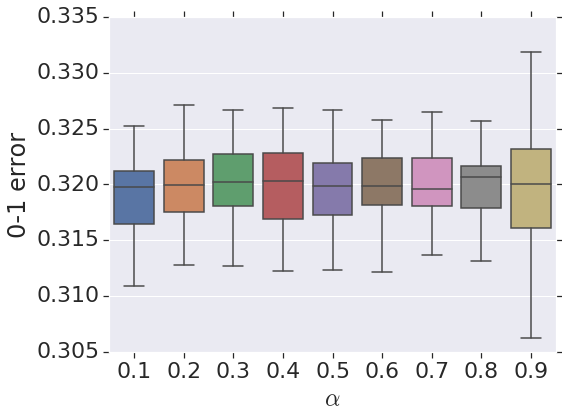}
	\quad
	\includegraphics[scale=0.27]{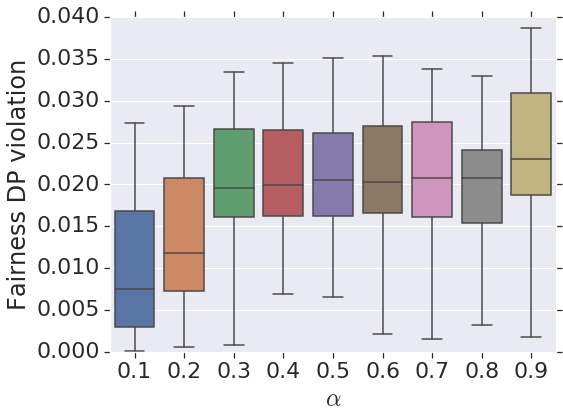}

	\caption{Results on {\tt synth} (top) and {\tt adult} (bottom) datasets.
	The left and right panel 
	show that as $\alpha$ is varied,
	CVaR-based optimisation generally results in a decrease in predictive accuracy and fairness violation.
	}
	\label{fig:additional-cvar-01}
\end{figure*}

\begin{figure*}[!p]
	\centering
	\includegraphics[scale=0.28]{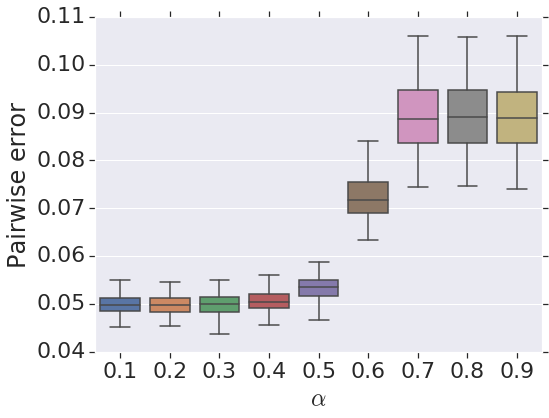}
	\quad
	\includegraphics[scale=0.28]{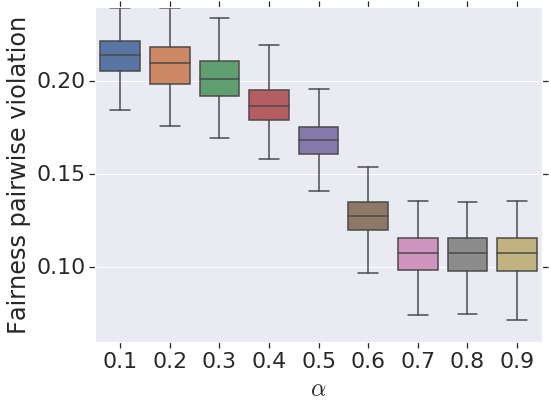}	

	\includegraphics[scale=0.27]{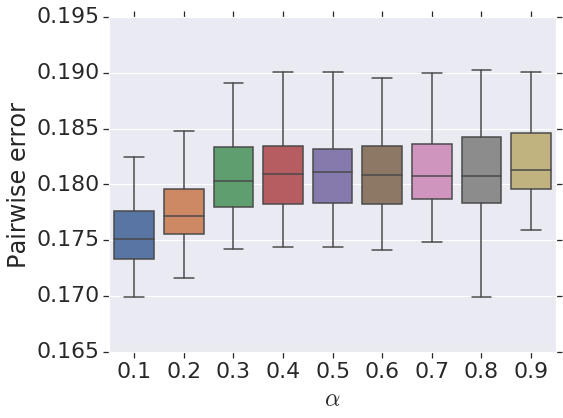}
	\quad
	\includegraphics[scale=0.27]{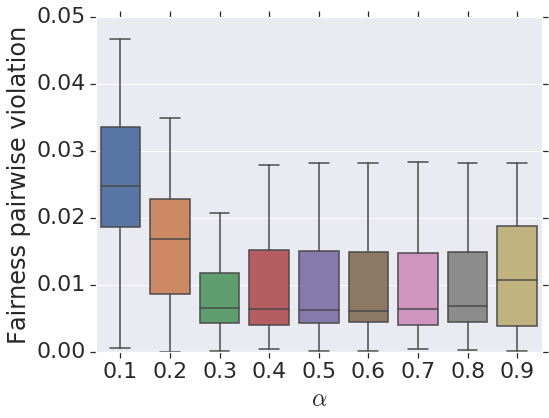}

	\caption{Results on {\tt synth} (top) and {\tt adult} (bottom) datasets.
	The left and right panel 
	show that as $\alpha$ is varied,
	CVaR-based optimisation generally results in a decrease in predictive accuracy and fairness violation.
	}
	\label{fig:additional-cvar-pd}
\end{figure*}

\begin{figure*}[!p]
	\centering
	\includegraphics[scale=0.28]{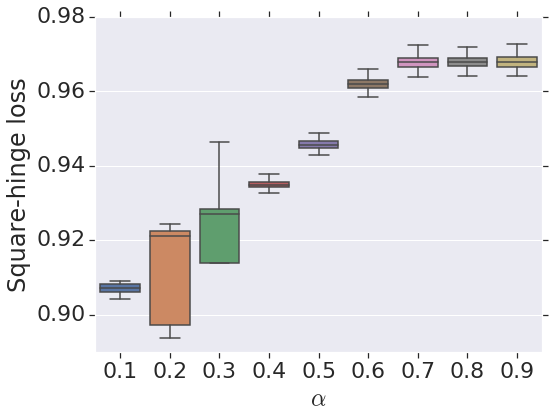}
	\quad
	\includegraphics[scale=0.28]{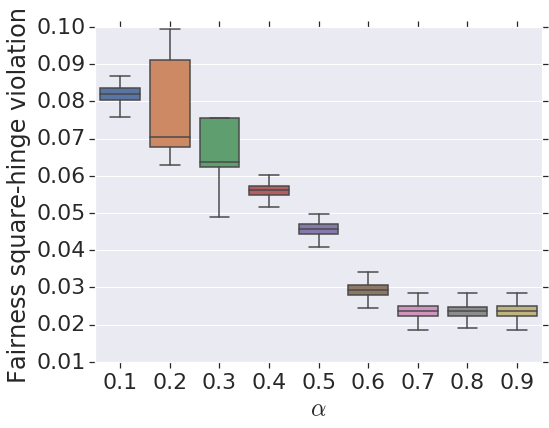}	

	\includegraphics[scale=0.27]{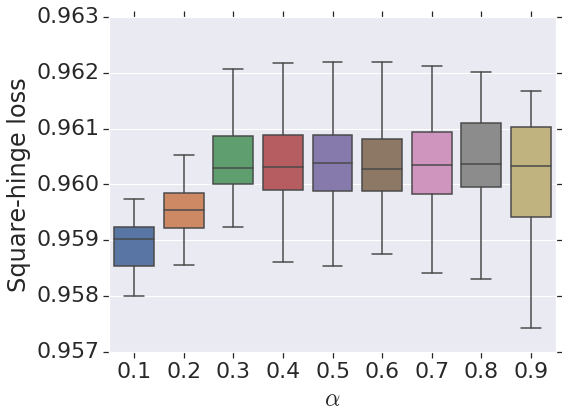}
	\quad
	\includegraphics[scale=0.27]{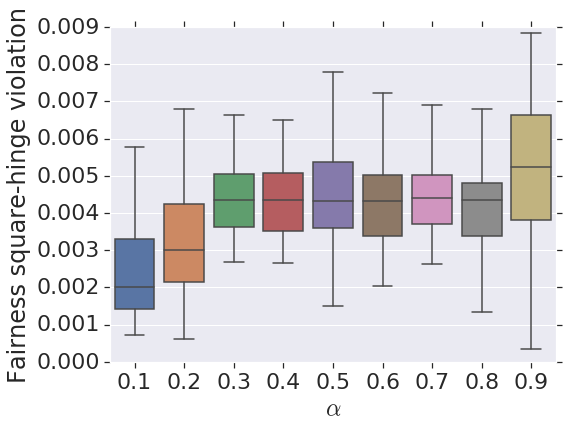}

	\caption{Results on {\tt synth} (top) and {\tt adult} (bottom) datasets.
	The left and right panel 
	show that as $\alpha$ is varied,
	CVaR-based optimisation generally results in a decrease in predictive accuracy and fairness violation.
	}
	\label{fig:additional-cvar-L2}
\end{figure*}

In Figures~\ref{fig:additional-cvar-pd} and \ref{fig:additional-cvar-L2}, we show the behaviour of the CVaR method as $\alpha$ is varied with respect to different metrics.
In Figure~\ref{fig:additional-cvar-01}, we measure the 0-1 error with respect to $\Ysf$,
and the difference of the 0-1 across the subgroups induced by $\Ssf$, i.e., the violation of the demographic parity (DP) condition.
Note that since the datasets are slightly imbalanced, the 0-1 error is not ideal as a measure of performance.

To better reflect the nature of class imbalance,
in Figure~\ref{fig:additional-cvar-pd}, we measure the pairwise disagreement (i.e., one minus the area under the ROC curve) with respect to $\Ysf$,
and the difference of the pairwise disagreement across the subgroups induced by $\Ssf$.
In Figure~\ref{fig:additional-cvar-L2}, we measure the square hinge loss with respect to $\Ysf$,
and the difference of this loss across the subgroups induced by $\Ssf$.
We generally see that, as in the body, increasing $\alpha$ has the effect of reducing predictive accuracy of $\Ysf$ while also reducing the fairness violation.

\end{document}